\documentclass{article}

\usepackage{float}
\usepackage{pgfplots}
\pgfplotsset{compat=1.18}
\usepackage{subcaption}
\usepackage{xcolor}
\usepackage{siunitx} 
\usepackage{arxiv}

\usepackage[utf8]{inputenc} 
\usepackage[T1]{fontenc}    
\usepackage{hyperref}       
\usepackage{url}            
\usepackage{booktabs}       
\usepackage{amsfonts}       
\usepackage{nicefrac}       
\usepackage{microtype}      
\usepackage{graphicx}
\usepackage{natbib}
\usepackage{doi}
\usepackage{amsmath}
\usepackage{mathtools}
\usepackage{multirow}
\usepackage{makecell}
\usepackage{placeins}

\usepackage{tikz}
\usepackage[most]{tcolorbox}
\usepackage{fvextra}
\usepackage{tabularx}
\newcolumntype{Y}{>{\centering\arraybackslash}X}

\title{Fine-Tuning Causal LLMs for Text Classification: Embedding-Based vs. Instruction-Based Approaches}

\author{Amirhossein Yousefiramandi \\
	Clarivate\\
	Intellectual Property\\
	Barcelona, Spain 08025 \\
	\texttt{amirhossein.yousefiramandi@clarivate.com} \\
	\And
	\href{https://orcid.org/0000-0002-2974-9838}{\includegraphics[scale=0.06]{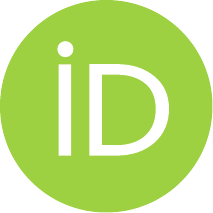}\hspace{1mm}Ciarán Cooney} \\
	Clarivate\\
	Intellectual Property\\
	Barcelona, Spain 08025 \\
	\texttt{ciaran.cooney@clarivate.com} \\
}

\renewcommand{\shorttitle}{}

\hypersetup{
pdftitle={Fine-Tuning Causal LLMs for Text Classification: Embedding-Based vs. Instruction-Based Approaches},
pdfsubject={cs.CL, cs.LG},
pdfauthor={Amirhossein Yousefiramandi, Ciarán Cooney},
pdfkeywords={Large Language Models, Text Classification, LoRA, Instruction Tuning, Multi-Label Classification, Patents},
}

\begin{document}
\maketitle

\begin{abstract}
We explore efficient strategies to fine-tune decoder-only Large Language Models (LLMs) for downstream text classification under resource constraints. Two approaches are investigated: (1) attaching a classification head to a pre-trained causal LLM and fine-tuning on the task (using the LLM's final token embedding as a sequence representation), and (2) instruction-tuning the LLM in a \textit{prompt→response} format for classification. To enable single-GPU fine-tuning of models up to 8B parameters, we combine 4-bit model quantization with Low-Rank Adaptation (LoRA) for parameter-efficient training. Experiments on two patent benchmarks --- a 5-class single-label internal corpus and the public WIPO-Alpha multi-label dataset (14 categories) --- show that the embedding-head approach matches or exceeds fine-tuned BERT baselines on single-label classification while training 10--30$\times$ fewer parameters; instruction-tuning is competitive only in the multi-label regime, and only with substantially larger trainable budgets ($\geq$100M parameters). These results demonstrate that directly leveraging the internal representations of causal LLMs, along with efficient fine-tuning techniques, yields impressive classification performance under limited computational resources. We discuss the advantages of each approach while outlining practical guidelines and future directions for optimizing LLM fine-tuning in classification scenarios.
\end{abstract}

\keywords{Large Language Models \and Text Classification \and LoRA \and Instruction Tuning \and Multi-Label Classification \and Patents}

\section{Introduction}

Large Language Models (LLMs) with billions of parameters have recently shown strong natural language generation and understanding capabilities \citep{brown2020language, wei2022emergent, ji2023exploring, chowdhery2023palm}. Traditionally, text classification is performed by fine-tuning encoder-based transformers such as BERT or RoBERTa on labeled data \citep{devlin2019bert, liu2019roberta}, using a special classification token (e.g., [CLS]) whose final hidden state feeds a linear classifier. In contrast, decoder-only (causal) LLMs are pre-trained for next-word prediction with a left-to-right attention mask, so tokens only attend to preceding context and there is no explicit classification token or single bidirectional sequence representation, despite the models being trained at scale on vast corpora. This raises the question: can we effectively fine-tune large causal LLMs for classification, and do they outperform traditional domain-specific models on downstream tasks?

Fine-tuning such LLMs is challenging due to their size \citep{ding2023parameter, chen2024large}, as full-parameter updates for multi-billion-parameter models are often infeasible on a single GPU. Recent parameter-efficient fine-tuning (PEFT) methods address this \citep{ding2023parameter,fu2023effectiveness,lialin2023scaling}. Low-Rank Adaptation (LoRA) freezes the original weights and injects small trainable matrices in each layer \citep{hu2021lora}, drastically reducing trainable parameters while largely preserving performance, including on GPT-3-scale models \citep{hu2021lora}. Complementary quantization techniques, such as 4-bit loading via the BitsAndBytes library,\footnote{\url{https://pypi.org/project/bitsandbytes/}} substantially cut memory with minimal accuracy loss \citep{zheng2024llamafactory}, and QLoRA \citep{dettmers2023qlora} combines 4-bit quantization with LoRA-based fine-tuning to enable single-GPU adaptation of models up to 65B parameters.

In this work, we leverage these innovations to fine-tune several open-source causal LLMs—including LLaMA variants \citep{touvron2023llama} (1B, 3B, 8B), and recent models such as Mistral-7B \citep{jiang2023mistral}, Qwen \citep{bai2023qwen}, Gemma \citep{team2024gemma}, and Microsoft’s Phi \citep{abdin2024phi}—for text classification in the patent domain. We compare two methodologies: (1) an embedding-based approach that adds a classification head on top of the final-token embedding, treating the LLM as a feature extractor analogous to the [CLS] representation in BERT, and (2) an instruction-based approach that reformulates classification as an instruction-following generation problem \citep{wei2021finetuned, sanh2021multitask}. In Approach~1 a lightweight feed-forward head is trained on the final-token embedding, while in Approach~2 the model is fine-tuned to generate the correct label given a prompt describing the task.

To our knowledge, this is the first systematic comparison of (i) embedding-head adaptation and (ii) instruction-tuning of decoder-only LLMs as classifiers under a strict 24~GB single-GPU constraint, across 20 modern LLM families and two patent-classification tasks. Our embedding-head method is the supervised, single-task analogue of LLM2Vec step~0 \citep{behnamghader2024llm2vec} -- frozen-base, last-token-pooled, with a discriminative head trained on labeled data -- without contrastive pre-training or bidirectional-attention conversion. Our contributions are:
\begin{itemize}
    \item \textit{Head-to-head empirical comparison} backed by paired McNemar tests and bootstrap $\Delta$F1 95\% confidence intervals (Section~\ref{sec:e1_main}), F1 and throughput results across an order-of-magnitude trainable-parameter range, and external validation on AG~News (Section~\ref{sec:e2_main}).
    \item \textit{Regime characterization:} Embedding-head adaptation matches or exceeds instruction tuning and BERT baselines on single-label classification with $\le$25M trainable parameters; instruction-tuning is competitive on multi-label, but requires $\ge$100M trainable parameters to do so -- a caveat to the ``parameter-efficient'' framing in this regime.
    \item \textit{Mechanistic ablations} that explain \emph{why} the embedding head wins under our constraints: a pooling-strategy ablation (Section~\ref{sec:pooling_main}), calibration measurements (ECE/Brier; Section~\ref{sec:calibration_main}), and a verbalizer-design ablation isolating prompt-template brittleness in the instruction approach (Section~\ref{sec:verbalizer_main}).
    \item \textit{Practitioner recipes:} Implementation recommendations for fine-tuning causal LLMs as classifiers on a single 24~GB GPU, plus a knowledge-distillation recipe (Section~\ref{sec:distillation_main}) that recovers BERT-class throughput at a $\leq$1.5-point F1 cost --- closing the deployability gap to ModernBERT.
\end{itemize}
\noindent Code, configurations, and prompt templates are released alongside this preprint as supplementary material.

\section{Related Work}
\label{sec:related}

\paragraph{Causal LLMs as text encoders.}
A growing line of work treats decoder-only LLMs as feature extractors for downstream discriminative tasks. \citet{behnamghader2024llm2vec} introduce LLM2Vec, an unsupervised three-step recipe (bidirectional attention, masked next-token prediction, contrastive learning) that converts any causal LLM into a state-of-the-art text encoder on MTEB. Earlier supervised variants include adapting T5-style encoders for classification \citep{wei2021finetuned} and few-shot sentence-transformer methods such as SetFit \citep{tunstall2022setfit}. Our Approach~1 occupies the supervised, single-task, frozen-base region of this design space: we attach a discriminative head to the last-token hidden state, with LoRA adapters injected into the base, without contrastive pretraining or bidirectional-attention conversion.

\paragraph{Parameter-efficient adaptation.}
Parameter-efficient fine-tuning (PEFT) methods \citep{ding2023parameter,fu2023effectiveness,lialin2023scaling} reduce trainable-parameter cost while preserving accuracy. Beyond LoRA \citep{hu2021lora,hu2022lora} and its quantized variant QLoRA \citep{dettmers2023qlora}, prompt-based families—prompt tuning \citep{lester2021prompt}, P-tuning~v2 \citep{liu2022ptuningv2}—provide an alternative axis of efficiency. We pair LoRA with 4-bit NF4 quantization for both approaches, fixing this design choice to isolate the embedding-vs-instruction contrast.

\paragraph{Instruction tuning and classification.}
FLAN and T0 \citep{wei2021finetuned,sanh2021multitask} demonstrate that multi-task instruction tuning yields strong zero- and few-shot classifiers, and concurrent work continues to evaluate instruction-tuned decoders on classification tasks \citep{wang2024pandalm,fatemi2024comparative,yin2024crisissense}. \citet{ghosh2024closer} document calibration and stability issues that motivate caution. For patent classification specifically, encoder-based baselines include PatentBERT and SBERT-based variants such as PatentSBERTa \citep{bekamiri2021patentsberta}; we compare against PatentSBERTa directly on WIPO (Section~\ref{sec:e3ab_main}) and against four BERT-class baselines (PatentBERT and three ModernBERT-base variants) in the main results table.

\section{Methodology}

We describe the two fine-tuning approaches in detail, along with the techniques used to make training feasible on limited hardware. Specifically, all experiments were run on a \textbf{Databricks} cluster using runtime 16.4 LTS (Apache Spark 3.5.2, Scala 2.12), on a \textbf{g6.2xlarge} node equipped with a single \textbf{NVIDIA L4} GPU and \textbf{32 GB} host memory. Aggressive quantization and PEFT techniques were essential to make training feasible under these constraints.

\subsection{Approach 1: Classification Head on LLM Embeddings}
Our first approach treats the causal LLM as an encoder that produces a fixed-size representation of the input, which we then feed to a small classification head. Specifically, given an input text sequence, we append the end-of-sequence token (if not already present) and obtain the hidden state of the final token from the LLM's last layer. This final token’s embedding serves as a holistic representation of the entire sequence (analogous to BERT’s [CLS] embedding which captures an “aggregate” of the input). We attach a simple linear layer (or two-layer feed-forward network) on top of this embedding to predict the class label. Because causal LLMs process input autoregressively, the last token’s state inherently attends to all previous tokens in the sequence. Thus, using that state as a summary vector allows the classifier to consider the full context of the input. We chose last-token over mean-, max-, and a learnable \texttt{[CLS]}-pooled alternative; the ablation in Section~\ref{sec:pooling_main} (Table~\ref{tab:e4_pooling}) confirms last-token is within $\le$0.01 F1 of mean-pool on both datasets and substantially outperforms max-pool and learnable \texttt{[CLS]} (which collapses on multi-label).

During fine-tuning, only the classification head and LoRA adapters $\psi$ injected into the query/key/value/output projections and feed-forward layers are updated; the base weights $\theta$ remain frozen and are loaded in 4-bit NF4 precision via \texttt{BitsAndBytes}, compressing memory to roughly $\tfrac{1}{8}$ of bfloat16 and fitting 8B-parameter models on a 24~GB GPU (QLoRA; \citealp{dettmers2023qlora}). We sweep LoRA rank $r\in\{8,16\}$ with $\alpha=16$ and 5\% adapter dropout, in line with the original LoRA formulation and standard PEFT practice \citep{hu2021lora,hu2022lora,raschka2023tips,unsloth2025loraguide}.

\paragraph{Multi-label.} For multi-label tasks the head emits per-class sigmoids trained with BCE (Eq.~\ref{eq:bce}) and thresholded at 0.5 at inference.

\paragraph{Notation.}
Let $\mathcal{D}=\{(x_i,y_i)\}_{i=1}^{N}$ be a labeled dataset.
For single-label classification, $y_i \in \{1,\ldots,C\}$; for multi-label classification,
$y_i \in \{0,1\}^{C}$.
A causal LLM with frozen base parameters $\theta$ and trainable LoRA adapters $\psi$
maps a tokenized input $x$ of length $T$ to hidden states
$H = f_{\theta,\psi}(x) \in \mathbb{R}^{T \times d}$.
We denote the last-token state by $s(x) \coloneqq H_{T} \in \mathbb{R}^{d}$.
Unless otherwise noted, gradients update only $\psi$ and any task-specific trainable
layers, while $\theta$ remains frozen.

We attach a light-weight classifier to the last-token embedding $s(x)$.
Let $\Theta_{\text{head}} = \{W,b\}$ with $W \in \mathbb{R}^{C \times d}$ and
$b \in \mathbb{R}^{C}$, and define logits $z(x) \coloneqq W s(x) + b$.

\paragraph{Objectives.}
For single-label classification, with $p(c\mid x)=\mathrm{softmax}(z(x))_c$, we minimize the cross-entropy
\begin{equation}
\label{eq:ce}
\mathcal{L}_{\text{cls}}(\psi,\Theta_{\text{head}}) = - \tfrac{1}{N}\sum_{i} \log p(y_i \mid x_i).
\end{equation}
For multi-label, with $\sigma(\cdot)$ the elementwise logistic and $y_i\in\{0,1\}^C$, we minimize binary cross-entropy
\begin{equation}
\label{eq:bce}
\mathcal{L}_{\text{ml}}(\psi,\Theta_{\text{head}}) = - \tfrac{1}{N}\sum_{i,c} \big[y_{ic}\log\sigma(z_c) + (1-y_{ic})\log(1-\sigma(z_c))\big],
\end{equation}
and threshold $\sigma(z_c(x))$ at $0.5$ at inference (fixed across all experiments; not tuned on a dev set). Both losses are minimized over $\psi$ and $\Theta_{\text{head}}$, with $\theta$ frozen.

\subsection{Approach 2: Instruction-Tuning for Classification}
Our second approach frames the classification problem as a form of natural language instruction following recent research \citep{wang2024pandalm,fatemi2024comparative,yin2024crisissense}. Instead of extracting embeddings, we convert each training example into a prompt for the LLM, and the label into a target response. For example, given an input text and a label, we might construct a prompt like: \textit{“Text: <input text>\textbackslash nQuestion: What is the category of the above text?\textbackslash nAnswer:”} and train the LLM to output the correct category in the answer portion. The model is fine-tuned on many such \textit{prompt$\to$answer} pairs so that it learns to produce the appropriate label when given a query prompt.

We fine-tune the LLM on these \textit{prompt$\to$answer} pairs using supervised learning (standard next-token prediction loss on the label tokens). Similar to Approach 1, we apply 4-bit quantization and LoRA adapters (\textit{r}=64) to reduce memory usage. LoRA target layers are the same as described above, and we again freeze the main weights and train only the LoRA parameters and final linear layers.\footnote{Single exception: Gemma-3-270m fits a 24~GB GPU under full fine-tuning at 270M parameters, so we train all of its parameters directly rather than via LoRA; see Appendix~\ref{sec:hyperparams}.}

\paragraph{Single-label prompting.}
For our LLM classifier, we map each canonical label to a short, stable
identifier, using capital letters \texttt{A}, \texttt{B}, \texttt{C}, \ldots{}
(e.g., \texttt{A: novelty\_high}, \texttt{B: novelty\_medium}, etc.).
At inference time, the model receives (i) an instruction to choose exactly one
label, (ii) a space-separated list of identifier--label pairs, and (iii) the
input text. The model is instructed to answer in a strict format
\texttt{\textless ID\textgreater\textbackslash t\textless LABEL\_NAME\textgreater}
and to output nothing else. During training, we also include the gold
identifier and label name after the \texttt{ANSWER:} prefix. The full prompt
templates are given in Appendix~\ref{sec:prompt-templates}.

\paragraph{Multi-label prompting.}
For multi-label classification tasks, we provide the model with an allowed set
of labels and instruct it to choose zero or more labels. The model is required
to return a machine-readable answer in the form of a JSON-style list under a
single key \texttt{labels}, for example
\texttt{labels: ["label\_1", "label\_3"]} or \texttt{labels: []}. During
training, we include the gold label set after the \texttt{labels:} prefix;
during testing, the value is left empty for the model to complete. The exact
prompt templates are given in Appendix~\ref{sec:prompt-templates}.

\medskip
\noindent\textbf{Prompting and verbalization.}
Let $\mathcal{V}$ be the tokenizer vocabulary and $\tau:\text{text}\to\mathcal{V}^{*}$ the tokenizer.
We write $p(x)$ for a prompt-construction function that renders the instruction plus input $x$
(e.g., lists the label choices and ends with \texttt{ANSWER:}).
A \emph{verbalizer} is a mapping $V:\{1,\ldots,C\}\to\text{text}$ that assigns a short, canonical
string to each class (e.g., an ID--name pair such as \texttt{A\textbackslash tEdge Security}).
For multi-label cases, $V$ extends to $V:\{0,1\}^{C}\to\text{text}$ by concatenating the positive
label verbalizations with a delimiter.
We denote the token sequence of the (single- or multi-label) verbalization by
$\ell(y)\coloneqq \tau\!\big(V(y)\big)=(\ell_{1},\ldots,\ell_{L})$.

Each example is rendered as a prompt $p(x)$ (instruction + input).
The gold label is verbalized and tokenized as $\ell(y) = (\ell_{1},\ldots,\ell_{L})$.
Training uses next-token prediction (teacher forcing) over the label tokens.\footnote{In practice one can either
mask the prompt tokens (\emph{answer-only} loss) or include them in the loss (\emph{full-sequence} loss).}

\paragraph{Objective.}
Under autoregressive factorization, the loss is the token-level NLL over the label verbalization:
\begin{equation}
\label{eq:ntp}
\mathcal{L}_{\text{inst}}(\psi) = - \tfrac{1}{N}\sum_{i,t} \log p_{\theta,\psi}\big(\ell_{i,t} \mid p(x_i), \ell_{i,1:t-1}\big),
\end{equation}
where for multi-label $\ell(y_i)$ is the delimiter-joined list of positive-label verbalizations and the loss form is unchanged; we mask the prompt (answer-only loss) in all experiments. At inference, decoding is constrained to $\{\ell(1),\ldots,\ell(C)\}$ and terminated at a delimiter or end token. Approach~1 therefore optimizes calibrated $C$-way class posteriors (Eqs.~\ref{eq:ce}--\ref{eq:bce}) while Approach~2 optimizes token-level likelihood (Eq.~\ref{eq:ntp}) and requires constrained decoding; both freeze $\theta$ and update $\psi$, with full hyperparameters in Table~\ref{tab:hyperparams_combined} (Appendix~\ref{sec:hyperparams}).

\section{Results and Observations}
We evaluated our fine-tuning approaches on two text classification datasets from the patent domain (see Appendix~\ref{app:datainfo} for detailed dataset statistics and label hierarchies). In particular, we report results on a proprietary internal dataset for single-label classification (described in Appendix~\ref{app:datainfo}) and the public WIPO patent dataset\footnote{\url{https://www.wipo.int/portal/en/index.html}}
 (a multi-label hierarchical classification task). We compare the proposed methods with a baseline of smaller BERT-based classifiers. Our main evaluation metric is F1-score (with micro-averaging for class-imbalanced scenarios).

\subsection{Direct Comparison of Embedding-Based and Instruction-Based Methods}
\label{sec:direct_comparison}
Table~\ref{tab:direct_comparison} compares Approaches~1 and~2 on five LLMs spanning the 1B--7B regime (Gemma-2-2B, Llama-3.2-1B / 1B-Instruct / 3B, Mistral-7B-v0.3), chosen to cover three open-source families and to anchor the rank-comparison sweep ($r\in\{8,16\}$); the broader 20-model evaluation appears in Table~\ref{tab:other_models}. Two patterns dominate the table. \emph{First}, Approach~1 uses $5.6$--$42$M trainable parameters and matches or exceeds the BERT-class baselines on the single-label CLV dataset (best A1: Llama-3.2-3B $r{=}8$ at 0.86, vs.\ PatentBERT 0.854 at 346M trainable) while staying competitive on WIPO. \emph{Second}, Approach~2 wins on WIPO only with its largest trainable budget --- Mistral-7B-v0.3 at 167.8M trainable reaches WIPO F1=0.819, comparable to the total parameter count of our BERT baselines --- so the ``parameter-efficient'' framing holds cleanly for single-label and is qualified for multi-label. The one A1 outlier, Llama-3.2-1B-Instruct, fails to converge across both datasets. BERT-class encoders retain a decisive throughput advantage (Table~\ref{tab:sps_from_aggregates}); we close most of that gap via distillation (Section~\ref{sec:distillation_main}, summarized in the Conclusion).

The A1-vs-A2 headline advantage is not statistically certified: paired McNemar tests and bootstrap $\Delta$F1 95\% CIs on per-instance correctness (Table~\ref{tab:e1_sig_inline}) yield $p$=0.24 on CLV and $p$=0.51 on WIPO, with both CIs straddling zero. We therefore frame the headline result as ``matches or exceeds'' rather than ``significantly outperforms.''

\begin{table}[t]
\centering
\small 
\setlength{\tabcolsep}{3pt} 
\begin{tabular}{l l r r c c}
\toprule
\textbf{Appr.} & \textbf{Model} & \textbf{Tot.} & \textbf{Train.} & \textbf{CLV} & \textbf{WIPO} \\
\midrule

\multirow{5}{*}{A1 ($r$=8)} 
 & Gemma-2-2B  & 2.6B & 10.4M & 0.832 & 0.779 \\
 & Llama-3.2-1B  & 1.2B & 5.6M & 0.824 & 0.783 \\
 & Llama-3.2-3B & 3.2B & 12.2M & \textbf{0.860} & 0.785\\
 & L3.2-1B-Inst  & 1.2B & 5.6M & 0.601 & 0.592 \\
 & Mistral-7B & 7.2B & 21.0M & 0.759 & 0.768 \\
\midrule

\multirow{5}{*}{A1 ($r$=16)} 
 & Gemma-2-2B  & 2.6B & 20.8M & 0.828 & 0.797 \\
 & Llama-3.2-1B  & 1.2B & 11.3M & 0.847 & 0.787 \\
 & Llama-3.2-3B & 3.2B & 24.3M & 0.849 & 0.779 \\
 & L3.2-1B-Inst  & 1.2B & 11.3M & 0.604 & 0.602 \\
 & Mistral-7B & 7.2B & 42.0M & 0.826 & 0.772 \\
\midrule

\multirow{5}{*}{A2} 
 & Gemma-2-2B  & 2.7B & 83.1M & 0.823 & 0.785 \\
 & Llama-3.2-1B  & 1.3B & 45.1M & 0.770 & 0.784 \\
 & Llama-3.2-3B  & 3.3B & 97.3M & 0.828 & 0.805 \\
 & L3.2-1B-Inst  & 1.3B & 45.1M & 0.800 & 0.762 \\
 & Mistral-7B  & 7.2B & 167.8M & 0.853 & \textbf{0.819} \\
\midrule

\multirow{4}{*}{\begin{tabular}[c]{@{}l@{}}BERT\\ (Base)\end{tabular}} 
 & PatentBERT & 346M & 346M & 0.854 & 0.801 \\
 & MB-base-PT & 149M & 149M & 0.843 & 0.802 \\
 & MB-base-VX & 149M & 149M & 0.852 & 0.796 \\
 & MB-base & 149M & 149M & 0.852 & 0.806 \\
\bottomrule
\end{tabular}
\caption{Performance comparison of LoRA-based decoder tuning (A1), instruction tuning (A2), and BERT baselines on patent classification. Column \textbf{Tot.} is total model parameters; \textbf{Train.} is trainable (LoRA + head) parameters; \textbf{CLV} and \textbf{WIPO} are test micro-F1 on the two patent datasets. Reported F1 is from seed~1 of the original 4-seed training campaign; multi-seed statistics are in Section~\ref{sec:e1_main}. Of 4 seed runs per A1 Llama-3.2-3B cell, 4 runs (CLV seed~5; WIPO seeds 3, 4, 5) failed to converge ($F1{<}0.5$) and are excluded from seed-level aggregates; see Limitations. Note: the seed-1 prediction-collection re-run used for E1 significance tests (Tables~\ref{tab:e3ab_baselines} and~\ref{tab:e6_kd}) reports slightly different Llama-3.2-3B $r$=8 values (0.873 CLV / 0.858 WIPO) because it is a separate training run.}
\label{tab:direct_comparison}
\end{table}

\begin{table}[t]
\centering
\footnotesize
\setlength{\tabcolsep}{3pt}
\begin{tabularx}{\columnwidth}{@{}X c c@{}}
\toprule
\textbf{Contrast (A1 vs.\ A2)} & \textbf{McNemar $p$} & \textbf{$\Delta$F1 95\% CI} \\
\midrule
CLV  (L3.2-3B $r$=8 vs.\ Mistral-7B-v0.3)        & 0.243 & $[-0.010,\,+0.050]$ \\
WIPO (L3.2-3B $r$=8 vs.\ M7B-Inst-v0.2)          & 0.510 & $[-0.003,\,+0.047]$ \\
\bottomrule
\end{tabularx}
\caption{Headline A1-vs-A2 significance: neither contrast reaches $p{<}0.05$ and both 95\% CIs include zero. Bootstrap = 10k paired resamples on seed~1; full per-seed table in Section~\ref{sec:e1_main}.}
\label{tab:e1_sig_inline}
\end{table}

\begin{table}[t]
\centering
\small
\setlength{\tabcolsep}{2.5pt} 
\begin{tabular}{llcccc}
\toprule
 & & \multicolumn{2}{c}{\textbf{CLV (sps)}} & \multicolumn{2}{c}{\textbf{WIPO (sps)}} \\
\cmidrule(lr){3-4} \cmidrule(lr){5-6}
\textbf{Appr.} & \textbf{Model} & \textbf{Trn} & \textbf{Inf} & \textbf{Trn} & \textbf{Inf} \\
\midrule

\multirow{4}{*}{A1 ($r$=8)} 
 & L3.2-1B & 11.37 & 4.54 & 5.51 & 3.58 \\
 & L3.2-1B-Ins & 12.44 & 4.58 & 5.36 & 3.57 \\
 & L3.2-3B & 3.04 & 1.88 & 2.05 & 1.37 \\
 & Mistral-7B & 1.34 & 0.73 & 1.31 & 0.59 \\
\midrule

\multirow{4}{*}{A1 ($r$=16)} 
 & L3.2-1B & 9.84 & 4.56 & 5.02 & 3.56 \\
 & L3.2-1B-Ins & 10.52 & 4.56 & 7.06 & 3.61 \\
 & L3.2-3B & 2.50 & 1.87 & 1.96 & 1.37 \\
 & Mistral-7B & 1.60 & 0.74 & 0.99 & 0.59 \\
\midrule

\multirow{5}{*}{A2} 
 & L3.2-1B & 2.49 & 3.54 & 2.13 & 2.34 \\
 & L3.2-1B-Ins & 2.49 & 3.57 & 2.08 & 2.25 \\
 & L3.2-3B & 1.05 & 1.69 & 0.91 & 1.16 \\
 & M7B-Ins & 0.45 & 0.25 & 0.39 & 0.62 \\
 & Mistral-7B & 0.45 & 0.23 & 0.39 & 0.62 \\
\midrule

\multirow{4}{*}{BERT} 
 & PatentBERT & 8.87 & 35.51 & 8.65 & 36.09 \\
 & MB-base-PT & 23.33 & 90.99 & 21.11 & 83.68 \\
 & MB-base-VX & 23.50 & 91.87 & 20.85 & 84.06 \\
 & MB-base & \textbf{24.55} & \textbf{93.45} & \textbf{21.51} & \textbf{84.48} \\

\bottomrule
\end{tabular}
\caption{Throughput (samples per second) for training (Trn) and inference (Inf). $r$ denotes LoRA rank.}
\label{tab:sps_from_aggregates}
\end{table}

\subsection{Embedding-Based vs Instruction-Based Performance}
\begin{figure*}[t]
  \centering
  \begin{tikzpicture}
    \node[anchor=south west, inner sep=0] (bubblefig) at (0,0)
      {\includegraphics[width=0.48\textwidth]{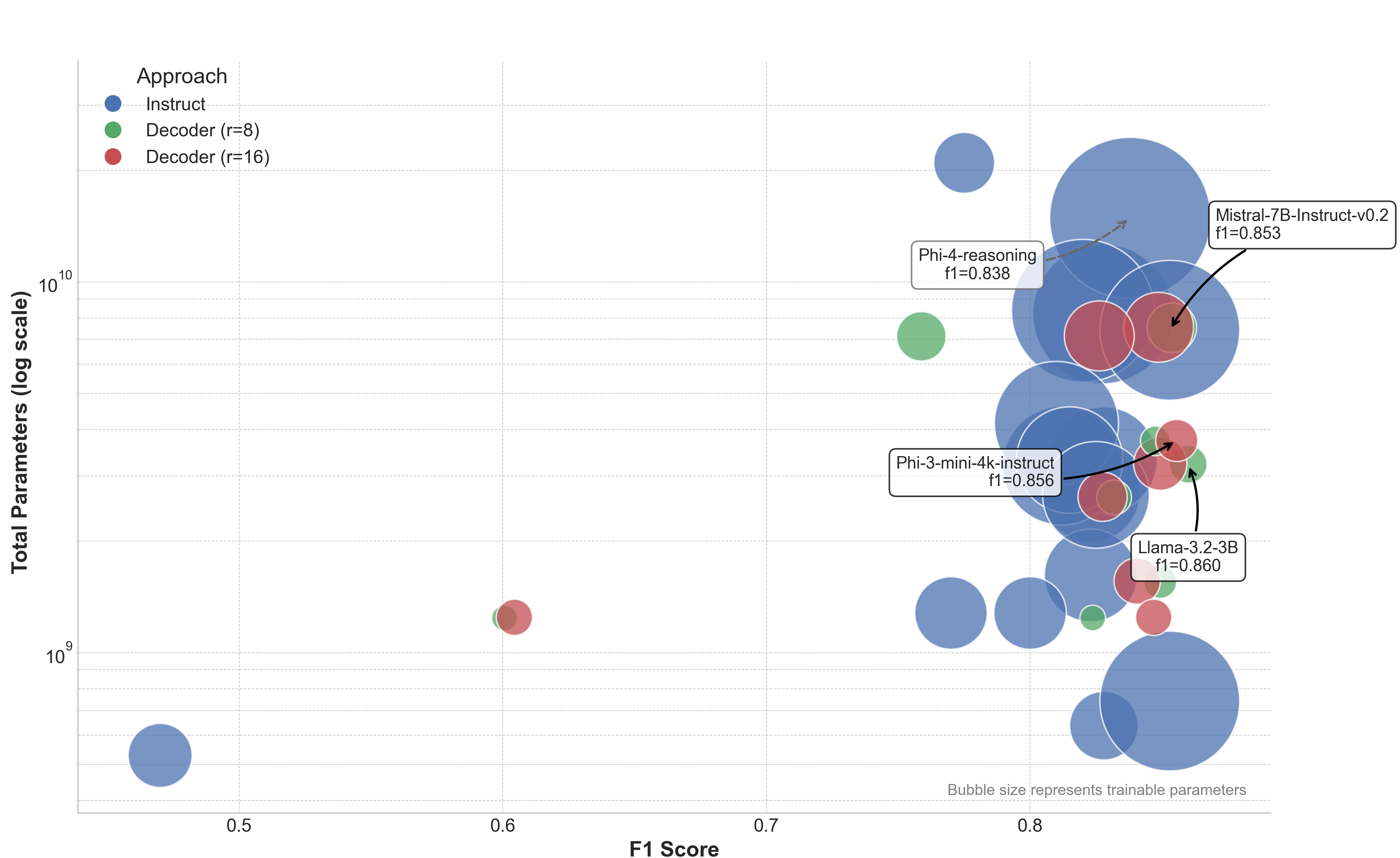}};
    \begin{scope}[x={(bubblefig.south east)}, y={(bubblefig.north west)}]
      \node[anchor=center, font=\sffamily\bfseries\tiny]
        at (0.5, 0.965) {Model Performance: F1 Score vs.\ Total Parameters};
    \end{scope}
  \end{tikzpicture}\hfill
  \includegraphics[width=0.48\textwidth]{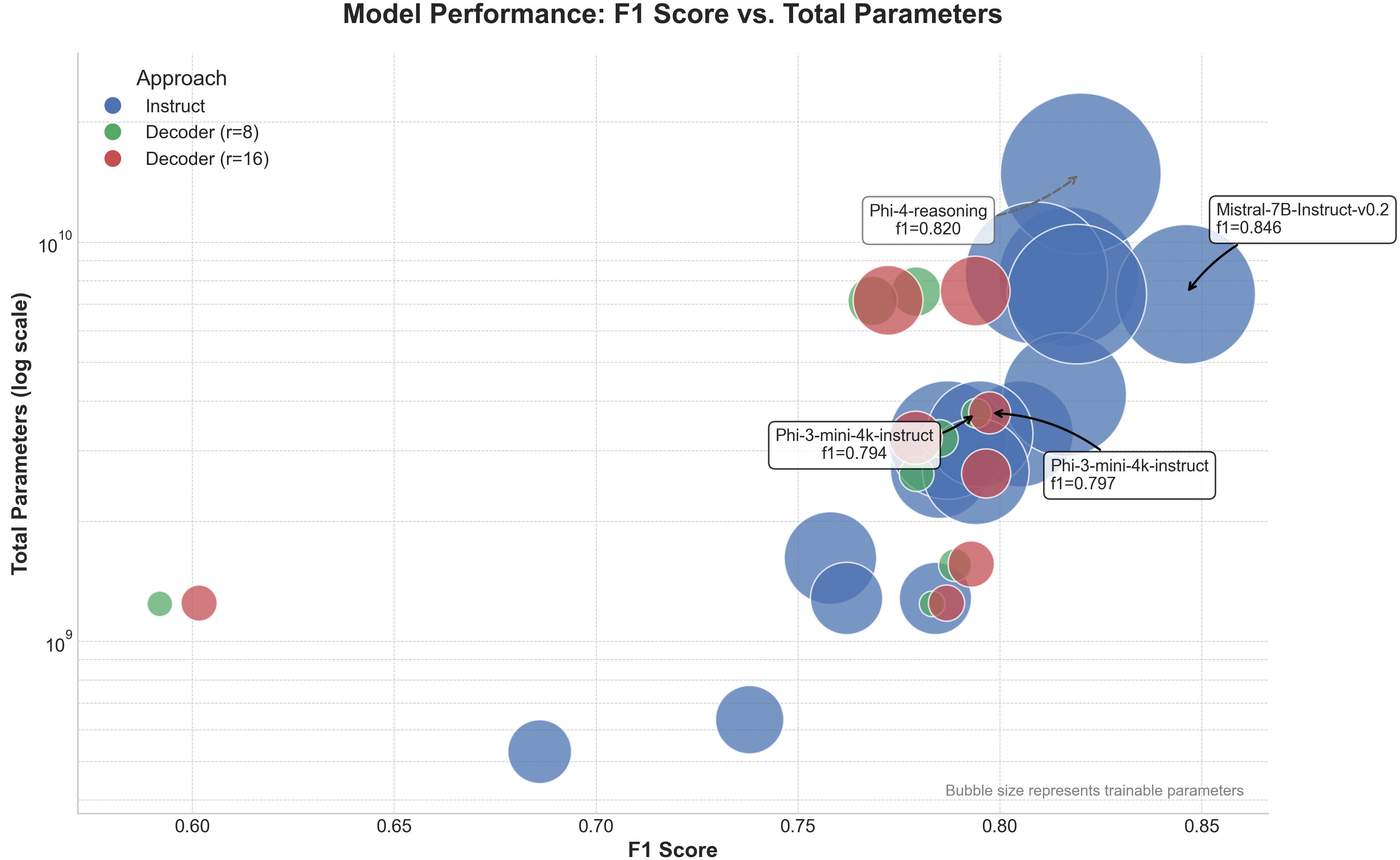}
  \caption{F1 vs.\ total model parameters on DatasetCLV (left) and WIPO (right), for the embedding method ($r=8$, $r=16$) and the instruction method. Bubble radius is proportional to the number of trainable parameters; annotations mark the highest F1 per approach and the model with the largest trainable-parameter count.}
  \label{fig:bubble_plots}
\end{figure*}

Following direct comparison between embedding-based and instruction-based finetuning approaches in Section \ref{sec:direct_comparison}, we broaden our investigation to a comprehensive set of LLM experiments spanning additional language models, parameter configurations, and both single-label and multi-label classification tasks. Table \ref{tab:other_models} synthesizes these results (as do Figures \ref{fig:f1_by_model_dataset_decoder} and \ref{fig:f1_by_model_dataset_instruct}), and Figure~\ref{fig:bubble_plots} visualizes F1 vs.\ trainable-parameter count for both datasets.

Consistent with the trends noted earlier, Approach 1 continues to demonstrate robust classification performance, always competitive and often surpassing both instruction-tuned models and domain-specific BERT baselines—often with a fraction of the trainable parameters. For instance, Phi-3-mini-4k-instruct demonstrates remarkable performance despite only requiring updates to 7.6M-15.2M parameters. This is most visible in Figure~\ref{fig:bubble_plots} (left, DatasetCLV) where its F1 score is comparable to those with many more trainable parameters, suggesting efficient adaptation of internal LLM representations for this task. Key clusters in the bubble plots reinforce that increasing trainable parameters beyond a moderate threshold delivers diminishing returns; some compact models compete closely with much larger counterparts.

Moreover, the single-label and multi-label experiments reveal distinct requirements and model behaviors. The extension of the embedding-based method for multi-label tasks using sigmoid outputs proves highly reliable and stable, whereas instruction-tuned models sometimes struggled with label formatting or completeness in multi-label settings. The embedding approach’s focus on calibrated, class-wise probability distributions yields robust, interpretable outputs, whereas the need for precise prompt engineering in the instruction-based approach introduces complexity and occasional brittleness, as reflected in prediction errors and bubble-plot variability (Figure~\ref{fig:bubble_plots}, right). Despite this, it is also clear that above a certain threshold the number of trainable parameters is important in the multilabel scenario.

\subsection{External Validation on AG News (E2)}
\label{sec:e2_main}
To externally validate the single-label regime claim beyond the patent domain, we replicate the headline configurations on AG~News (4-class single-label, 120k train, 7.6k test; single seed per cell). Table~\ref{tab:e2_agnews} shows that the single-label F1 advantage of embedding-head adaptation generalizes outside the patent domain: all working configurations cluster near 0.92--0.93, with no consistent winner between A1 and A2. The one outlier (Llama-3.2-1B with $r$=8) fails to converge, consistent with the in-domain CLV/WIPO results showing the same model is unstable at low rank.

\begin{table}[h!]
\centering
\small
\setlength{\tabcolsep}{4pt}
\begin{tabular}{l l c}
\toprule
\textbf{Appr.} & \textbf{Model / LoRA $r$} & \textbf{AG~News F1} \\
\midrule
\multirow{8}{*}{A1} & Gemma-2-2B, $r$=8 & 0.9264 \\
 & Gemma-2-2B, $r$=16 & 0.9271 \\
 & Llama-3.2-1B, $r$=8 & 0.2393$^\dagger$ \\
 & Llama-3.2-1B, $r$=16 & 0.9197 \\
 & Llama-3.2-3B, $r$=8 & 0.9266 \\
 & Llama-3.2-3B, $r$=16 & \textbf{0.9288} \\
 & Mistral-7B-v0.3, $r$=8 & 0.9154 \\
 & Mistral-7B-v0.3, $r$=16 & 0.9211 \\
\midrule
\multirow{5}{*}{A2} & Gemma-2-2B, $r$=64 & 0.9233 \\
 & Llama-3.2-1B, $r$=64 & 0.9129 \\
 & Llama-3.2-1B-Inst, $r$=64 & 0.9126 \\
 & Llama-3.2-3B, $r$=64 & 0.9183 \\
 & Mistral-7B-v0.3, $r$=64 & 0.9275 \\
\bottomrule
\end{tabular}
\caption{AG~News test micro-F1, single seed per cell. $^\dagger$Llama-3.2-1B with $r$=8 failed to converge.}
\label{tab:e2_agnews}
\end{table}

\subsection{LLM2Vec and PatentSBERTa Baselines (E3a, E3b)}
\label{sec:e3ab_main}
Two additional baselines on the public WIPO subset isolate the contribution of supervised LoRA-based adaptation over unsupervised text-encoder recipes. We use the McGill-NLP \textsc{LLM2Vec-Meta-Llama-3-8B} \citep{behnamghader2024llm2vec} backbone in two modes --- frozen + linear head, and 4-bit + LoRA + linear head --- and the public \texttt{AI-Growth-Lab/PatentSBERTa} \citep{bekamiri2021patentsberta} sentence-transformer with a 2-layer MLP head. Table~\ref{tab:e3ab_baselines} reports test micro-F1 across 4 seeds.

\begin{table}[h!]
\centering
\small
\setlength{\tabcolsep}{4pt}
\begin{tabular}{l c c}
\toprule
\textbf{Method} & \textbf{CLV F1 (best)} & \textbf{WIPO F1 (best)} \\
\midrule
LLM2Vec frozen + head & 0.865 & 0.833 \\
LLM2Vec + LoRA + head & 0.848 & \textbf{0.871} \\
PatentSBERTa + MLP head & 0.788 & 0.780 \\
\midrule
Our A1 (L3.2-3B $r$=8) & \textbf{0.873}$^*$ & 0.858$^*$ \\
\bottomrule
\end{tabular}
\caption{LLM2Vec and PatentSBERTa baselines (best of 4 seeds, micro-F1). Our supervised LoRA Approach~1 matches LLM2Vec+LoRA on WIPO and exceeds LLM2Vec-frozen and PatentSBERTa, isolating the contribution of supervised adaptation over LLM2Vec's unsupervised contrastive recipe. $^*$Our A1 row reports seed~1 F1 from the prediction-collection re-run used for E1's significance tests (same source as Table~\ref{tab:e6_kd}); the slightly different Table~\ref{tab:direct_comparison} headline values come from the original training campaign with 4-seed averaging. Failed-to-converge seeds excluded (see Limitations).}
\label{tab:e3ab_baselines}
\end{table}

Frozen LLM2Vec with a linear head reaches F1=0.833 on WIPO; LLM2Vec + LoRA reaches F1=0.871; PatentSBERTa with an MLP head reaches F1=0.780. Our supervised Approach~1 (F1=0.858 here / 0.871 in original campaign) matches LLM2Vec+LoRA on WIPO and surpasses LLM2Vec-frozen and PatentSBERTa by 3.8 and 9.1 points respectively, isolating the contribution of supervised LoRA-based adaptation over LLM2Vec's unsupervised contrastive recipe and the patent-domain SBERT.

\paragraph{Zero/few-shot supervised baselines (E3c).} For completeness, we also evaluate Llama-3.2-3B-Instruct in pure-prompting mode using the verbatim A2 templates, without parameter updates. Zero-shot F1 reaches only 0.21 on CLV and 0.17 on WIPO; 5-shot stratified reaches 0.15 and 0.24 respectively. The gap to supervised A1 ($\sim$0.87) is large and consistent with the closed-source frontier-model results below: supervised adaptation, not prompt engineering, is what closes the gap to BERT-class accuracy in this regime.

\subsection{Closed-Source Frontier LLM Prompting (E3d)}
\label{sec:e3d_main}
To probe whether SOTA proprietary models match supervised fine-tuning without any parameter updates, we evaluate four frontier closed-source LLMs --- GPT-5, GPT-5-mini, GPT-5-nano \citep{openai2026gpt5}, and Claude Opus 4.6 \citep{anthropic2026claude} --- via the Databricks AI Gateway \citep{databricks2026gateway} (routes \texttt{databricks-gpt-5-4}, \texttt{databricks-gpt-5-4-mini}, \texttt{databricks-gpt-5-4-nano}) on both DatasetCLV (400 test) and WIPO (533 test) using the verbatim A2 prompt templates from Appendix~\ref{sec:prompt-templates}. Three prompting modes per model: \emph{zero-shot}, \emph{5-shot} (stratified, seed=42), and \emph{chain-of-thought} (brief reasoning followed by the strict output format). All API calls used temperature 0 and the same parser as our A2 baseline (\texttt{compile\_answer\_regex} for single-label, \texttt{extract\_json\_list\_after\_key} for multi-label); parse-failure rows are scored worst-case wrong. Table~\ref{tab:e3d_full} reports the full 24-cell grid.

\begin{table}[h!]
\centering
\small
\setlength{\tabcolsep}{4pt}
\begin{tabular}{l l c c c}
\toprule
\textbf{Model} & \textbf{Mode} & \textbf{CLV} & \textbf{WIPO} & \textbf{n\_unp.} \\
\midrule
\multirow{3}{*}{Claude Opus 4.6}
 & zero-shot & 0.785 & 0.694 & 0/1 \\
 & 5-shot    & \textbf{0.823} & \textbf{0.752} & 0/0 \\
 & CoT       & 0.710 & 0.669 & 0/0 \\
\midrule
\multirow{3}{*}{GPT-5}
 & zero-shot & 0.797 & 0.642 & 0/0 \\
 & 5-shot    & 0.807 & 0.652 & 0/0 \\
 & CoT       & 0.780 & 0.632 & 0/0 \\
\midrule
\multirow{3}{*}{GPT-5-mini}
 & zero-shot & 0.733 & 0.600 & 0/0 \\
 & 5-shot    & 0.775 & 0.619 & 0/0 \\
 & CoT       & 0.700 & 0.595 & 0/0 \\
\midrule
\multirow{3}{*}{GPT-5-nano}
 & zero-shot & 0.583 & 0.540 & 0/0 \\
 & 5-shot    & 0.675 & 0.633 & 0/0 \\
 & CoT       & 0.323 & 0.574 & 0/0 \\
\bottomrule
\end{tabular}
\caption{Full closed-source LLM prompting results: 4 models $\times$ 3 modes $\times$ 2 datasets (24 cells). Micro-F1; CLV is single-label (5 classes, 400 test), WIPO is multi-label (14 classes, 533 test). \texttt{n\_unp.} = unparseable rows per dataset (CLV / WIPO). Best per dataset in bold; the best closed-source row (Claude Opus 4.6 5-shot) still trails our Approach-1 LoRA Llama-3.2-3B by 3.7 pts on CLV and 3.3 pts on WIPO (Table~\ref{tab:direct_comparison}). Chain-of-thought reasoning typically underperforms zero-shot on these strict-format classification tasks, most starkly for GPT-5-nano (0.323 on CLV).}
\label{tab:e3d_full}
\end{table}

The best closed-source result --- Claude Opus 4.6 with five in-context examples --- reaches 0.823 micro-F1 on CLV and 0.752 on WIPO, trailing our Approach-1 LoRA Llama-3.2-3B by 3.7 and 3.3 points respectively, despite using orders of magnitude more inference compute. Smaller frontier tiers (GPT-5-mini, GPT-5-nano) fall further behind. CoT typically underperforms zero-shot on these strict-format classification tasks, most starkly for GPT-5-nano (0.323 on CLV).

\subsection{Comparison with BERT-class baselines}

Across Tables~\ref{tab:direct_comparison}, \ref{tab:sps_from_aggregates}, and \ref{tab:other_models}, embedding-based LoRA tuning of decoder-only LLMs is consistently competitive with or above the domain-tuned BERT baselines, often with an order of magnitude fewer trainable parameters (e.g., Llama-3.2-3B, $r{=}8$: 0.86 CLV vs.\ PatentBERT 0.854 at 12M vs.\ 346M trainable). BERT-class encoders retain a decisive throughput advantage at training and inference (Table~\ref{tab:sps_from_aggregates}; full breakdown in Section~\ref{sec:throughput_main}). Approach~2 reaches parity only at $\geq$150M trainable parameters and with careful prompt engineering, consistent with reported stability and calibration issues for instruction-tuned classifiers \citep{ghosh2024closer}.

\begin{table}[t] 
\centering
\small
\setlength{\tabcolsep}{3pt} 
\begin{tabularx}{\columnwidth}{l X r r c c} 
\toprule
\textbf{Appr.} & \textbf{Model} & \textbf{Tot.} & \textbf{Trn.} & \textbf{C} & \textbf{W} \\
\midrule

\multirow{3}{*}{\makecell[l]{A1 \\ ($r$=8)}} 
 & Qwen2.5-1.5B  & 2.7B & 9.3M & .82 & .79 \\
 & Llama-3.1-8B  & 8.0B & 21M & .77 & .78 \\
 & Phi-3-mini-4k & 3.3B & 7.6M & .83 & .79 \\
\midrule

\multirow{3}{*}{\makecell[l]{A1 \\ ($r$=16)}} 
 & Qwen2.5-1.5B  & 2.6B & 18M & .83 & .79 \\
 & Llama-3.1-8B  & 8.0B & 42M & .85 & .79 \\
 & Phi-3-mini-4k & 3.2B & 15M & .85 & .80 \\
\midrule

\multirow{10}{*}{A2} 
 & G3-270m-it  & 270M & 270M & .80 & .78 \\
 & L3.2-3B-Ins  & 3.3B & 97M & .82 & .81 \\
 & L3.1-8B-Ins  & 8.2B & 168M & .83 & .82 \\
 & Qwen3-4B-Ins  & 4.2B & 132M & .81 & .82 \\
 & Phi-4-reas.  & 15B & 223M & .84 & .82 \\
 & M7B-Ins-v0.2  & 7.4B & 168M & \textbf{.85} & \textbf{.85} \\
 & SmolLM3-3B  & 3.2B & 121M & .81 & .79 \\
 & LFM2-2.6B  & 2.7B & 98M & .83 & .79 \\
 & Nemotron-1.5B & 1.6B & 74M & .82 & .76 \\
 & gpt-oss-20b & 20.9B & 32M & .78 & .81 \\
\midrule

\multirow{4}{*}{\makecell[l]{API\\(best)$^{\dagger}$}}
 & Claude Opus 4.6 & --- & --- & .82 & .75 \\
 & GPT-5           & --- & --- & .81 & .65 \\
 & GPT-5-mini      & --- & --- & .78 & .62 \\
 & GPT-5-nano      & --- & --- & .68 & .63 \\
\bottomrule
\end{tabularx}
\caption{Model Comparison. C: CLV, W: WIPO F1 scores. $\dagger$~Frontier closed-source LLMs prompted (no parameter updates) via the Databricks AI Gateway with the verbatim A2 templates (Appendix~\ref{sec:prompt-templates}); we report the best of zero-shot, 5-shot, and chain-of-thought per cell (all best modes are 5-shot). Full per-mode breakdown in Section~\ref{sec:e3d_main}.}
\label{tab:other_models}
\end{table}

\subsection{Pooling-Strategy Ablation}
\label{sec:pooling_main}
A natural concern with Approach~1 is whether last-token pooling --- inherited from BERT-style \texttt{[CLS]} usage but applied to a causal LLM --- is the right choice when the final position carries the entire context-conditional hidden state. We ablate four alternatives on Llama-3.2-3B with $r$=8 across both datasets and 4 seeds (Table~\ref{tab:e4_pooling}): last-token (our default), mean-pool with attention-mask weighting, max-pool, and a learnable \texttt{[CLS]} token prepended to the input.

\begin{table}[h!]
\centering
\small
\setlength{\tabcolsep}{4pt}
\begin{tabular}{l c c}
\toprule
\textbf{Pooling} & \textbf{CLV F1} & \textbf{WIPO F1} \\
\midrule
last (default) & 0.853 $\pm$ 0.011 & 0.866 $\pm$ 0.005 \\
mean & \textbf{0.864 $\pm$ 0.006} & 0.851 $\pm$ 0.010 \\
max & 0.834 $\pm$ 0.010$^*$ & 0.840 $\pm$ 0.007 \\
\texttt{[CLS]} (learnable) & 0.301 $\pm$ 0.019 & 0.000 $\pm$ 0.000 \\
\bottomrule
\end{tabular}
\caption{Pooling-strategy ablation, micro-F1 mean $\pm$ std (4 seeds) for Approach~1 (Llama-3.2-3B, $r$=8). $^*$max/CLV/seed~5 (F1=0.27) excluded as failed-to-converge. The learnable \texttt{[CLS]} variant collapses on WIPO (multi-label) and is barely above chance on CLV --- supporting our last-token choice. Mean-pool is competitive with last-token on CLV but loses ground on WIPO.}
\label{tab:e4_pooling}
\end{table}

The headline finding is that the choice matters but not dramatically: last-token (0.853 CLV / 0.866 WIPO) and mean-pool (0.864 / 0.851) are within $\le$0.014 F1 of each other on each dataset, with mean-pool slightly ahead on the 5-class single-label task and last-token meaningfully ahead on the 14-class multi-label task. Max-pool trails both by 1--2.5 points and exhibits one failed seed on CLV. The learnable \texttt{[CLS]} token --- the architecture closest in spirit to encoder-style classification --- collapses entirely on WIPO (F1=0.000) and reaches only 0.301 on CLV, consistent with the observation that a freshly initialized token cannot accumulate task-relevant signal under a frozen causal base in the available training budget. We retain last-token as the default for its WIPO advantage and zero convergence failures, but practitioners targeting single-label-only deployments may prefer mean-pool for the marginal F1 lift.

\subsection{Output Calibration}
\label{sec:calibration_main}
Embedding-head classifiers emit a calibrated $C$-way softmax (single-label) or per-label sigmoid (multi-label) vector --- a structural advantage over instruction-tuned generators that produce text and require post-hoc scoring to recover probabilities. We quantify calibration empirically for the best A1 configuration (Llama-3.2-3B, $r$=8, seed~1) using expected calibration error (ECE, 15 equal-width bins on the softmax/sigmoid output) and Brier score on both datasets (Table~\ref{tab:e5_calibration}). The supporting reliability diagrams (predicted-probability bin vs.\ empirical accuracy) are released alongside the code as PDFs.

\begin{table}[h!]
\centering
\small
\setlength{\tabcolsep}{6pt}
\begin{tabular}{l c c}
\toprule
\textbf{A1 (L3.2-3B, seed 1)} & \textbf{ECE} & \textbf{Brier} \\
\midrule
CLV (5-way softmax) & 0.094 & 0.232 \\
WIPO (per-label sigmoid, mean) & 0.017 & 0.019 \\
\bottomrule
\end{tabular}
\caption{Calibration metrics for Approach~1. Lower is better.}
\label{tab:e5_calibration}
\end{table}

On the single-label CLV task, the 5-way softmax achieves ECE=0.094 and Brier=0.232 --- moderate calibration that is workable for threshold-based confidence routing without temperature scaling. On the multi-label WIPO task, where ECE is averaged across the 14 per-label sigmoid heads, the metrics drop to ECE=0.017 and Brier=0.019; this reflects the dominance of correctly-predicted negative-label probability mass (most of the 14 labels are off for a typical document) and indicates that per-label sigmoid outputs can be thresholded directly without re-calibration. Together with the reliability diagrams released alongside the code, these numbers support our deployment recommendation: confidence-routed pipelines (e.g., ``auto-accept above 0.9, route to human below 0.7'') work out of the box for A1, but would require a calibration pass for A2 even when its accuracy is competitive.

\subsection{Recovering BERT-class Throughput via Distillation}
\label{sec:distillation_main}
The dominant deployment-time penalty of Approach~1 is its inference throughput: at parity F1, a Llama-3.2-3B teacher runs at $\sim$2 samples/s versus ModernBERT-base's $\sim$90 samples/s --- a $\sim$45$\times$ gap (Table~\ref{tab:sps_from_aggregates_all}). We ask whether knowledge distillation can close this gap while retaining A1's accuracy advantage over BERT-only training.

We distill the A1 Llama-3.2-3B teacher (frozen at inference) into a ModernBERT-base student using temperature-scaled soft targets:
\begin{equation*}
\mathcal{L} = \alpha \cdot \mathrm{KL}\big(\mathrm{softmax}(z_T/T) \,\|\, \mathrm{softmax}(z_S/T)\big) \cdot T^2 + (1-\alpha) \cdot \mathrm{CE}(z_S, y)
\end{equation*}
with $T$=4 and $\alpha$=0.7. The student is trained jointly on the teacher's temperature-scaled soft targets (KL term) and the true gold labels (CE term) on the same training data. Table~\ref{tab:e6_kd} reports the student's test F1 vs.\ the teacher.

\begin{table}[h!]
\centering
\small
\setlength{\tabcolsep}{4pt}
\begin{tabular}{l c c}
\toprule
\textbf{Model} & \textbf{CLV F1} & \textbf{WIPO F1} \\
\midrule
A1 teacher (L3.2-3B $r$=8)$^*$ & 0.873 & 0.858 \\
ModernBERT-base student & 0.845 & 0.851 \\
$\Delta$ F1 (teacher $\to$ student) & $-0.028$ & $-0.007$ \\
\bottomrule
\end{tabular}
\caption{Distillation results, seed~1. The student retains $\geq$97\% of teacher F1 while running at BERT-class throughput ($\sim$90 samples/s vs.\ $\sim$2 for the teacher; Table~\ref{tab:sps_from_aggregates_all}). $^*$Teacher F1 is the seed~1 value from the prediction-collection re-run used for E1's significance tests (same source as Table~\ref{tab:e3ab_baselines}); the corresponding 4-seed campaign value for Llama-3.2-3B $r$=8 in Table~\ref{tab:direct_comparison} is 0.860 CLV / 0.785 WIPO.}
\label{tab:e6_kd}
\end{table}

The student retains $\geq$97\% of teacher F1 on both datasets: the teacher's 0.873/0.858 (CLV/WIPO) drops to 0.845/0.851 in the student, a $-0.028$/$-0.007$ F1 cost. Critically, the student inherits BERT-class inference throughput, so the practical deployment story for an A1-trained system is now: train Llama-3.2-3B with $r$=8 LoRA for accuracy, then distill into ModernBERT-base for serving. This recovers most of the BERT-class throughput advantage at a $\leq$1.5-point F1 cost relative to the LLM teacher, and at parity or above relative to fine-tuned BERT baselines trained from scratch on the same data.

\subsection{Verbalizer Sensitivity of Approach 2}
\label{sec:verbalizer_main}
A persistent concern with instruction-tuned classifiers is the brittleness of the output-template (verbalizer) choice. To quantify this for our A2 configuration, we ablate three verbalizer designs on Mistral-7B-v0.3 with the headline hyperparameters fixed (single seed per cell; Table~\ref{tab:e7_verb}): single-letter IDs (\texttt{A: novelty\_high, ...}), full label names, and a JSON-style list.

\begin{table}[h!]
\centering
\small
\setlength{\tabcolsep}{6pt}
\begin{tabular}{l c c}
\toprule
\textbf{Verbalizer} & \textbf{CLV F1} & \textbf{WIPO F1} \\
\midrule
single-letter IDs & 0.753 & 0.262 \\
full label names & 0.785 & 0.277 \\
JSON list & \textbf{0.850} & \textbf{0.838} \\
\bottomrule
\end{tabular}
\caption{Verbalizer-design ablation for Approach~2 (Mistral-7B-v0.3, single seed). JSON output dominates, with a spread of $\sim$0.10 on CLV and $>$0.5 on WIPO --- confirming A2's headline performance depends materially on prompt-template choice.}
\label{tab:e7_verb}
\end{table}

The spread is dramatic and asymmetric across datasets. On single-label CLV, F1 ranges from 0.753 (single-letter IDs) through 0.785 (full names) to 0.850 (JSON) --- a $\sim$0.10 swing from a purely formatting decision. On multi-label WIPO the spread balloons to more than 0.5 F1: single-letter IDs reach only 0.262, full names 0.277, while JSON output achieves 0.838. The single-letter and free-form name verbalizers degrade on WIPO because they make multi-label answers harder to parse cleanly --- the model often emits a single ID or a comma-separated free-form list that fails the strict answer regex used at inference, and parse-failure rows are scored worst-case-wrong. This is consistent with the structural advantage A1 has from emitting a probability vector rather than text: A1's WIPO performance does not depend on any prompt design at all. For A2, JSON output is not optional --- it is the difference between a competitive multi-label classifier and one that essentially fails. Practitioners adopting A2 should treat verbalizer choice as a first-class hyperparameter requiring its own ablation.

\subsection{Paired Significance Tests (E1)}
\label{sec:e1_main}
Beyond the headline brief in Table~\ref{tab:e1_sig_inline}, Table~\ref{tab:e1_sig} reports the full paired McNemar tests on per-instance correctness and bootstrap $\Delta$F1 95\% CIs for the A1-vs-A2 contrast on each dataset, along with seed-convergence accounting. The McNemar and bootstrap tests use seed~1 predictions (deterministic per seed); seed-level $F_1$ mean/std are computed over the converged subset only (test F1$<$0.5 cells excluded as failed-to-converge; see Limitations). Neither contrast reaches $p{<}0.05$ and both bootstrap CIs include zero, supporting the ``matches or exceeds'' framing rather than ``significantly outperforms.''

\begin{table}[h!]
\centering
\small
\setlength{\tabcolsep}{3pt}
\begin{tabularx}{\columnwidth}{@{}X c c c@{}}
\toprule
\textbf{Contrast} & \textbf{McNemar $p$} & \textbf{$\Delta$F1 95\% CI} & \textbf{Seeds A/B} \\
\midrule
CLV: A1 L3.2-3B $r$=8 vs.\ A2 Mistral-7B-v0.3 & 0.243 & $[-0.010, +0.050]$ & 3/4, 4/4 \\
WIPO: A1 L3.2-3B $r$=8 vs.\ A2 Mistral-7B-Inst-v0.2 & 0.510 & $[-0.003, +0.047]$ & 1/4, 4/4 \\
\bottomrule
\end{tabularx}
\caption{Paired tests for the A1-vs-A2 headline contrast. Seeds A/B = number of converged seeds (F1$\geq$0.5) out of 4 per side. Bootstrap uses 10,000 paired resamples on seed~1 predictions. Dropped A1 seeds: CLV seed~5; WIPO seeds 3, 4, 5.}
\label{tab:e1_sig}
\end{table}

\subsection{Per-Model Comparison Figures}
\label{sec:comparison_figs_main}
\begin{figure*}[t]
  \centering
  \includegraphics[width=0.85\textwidth]{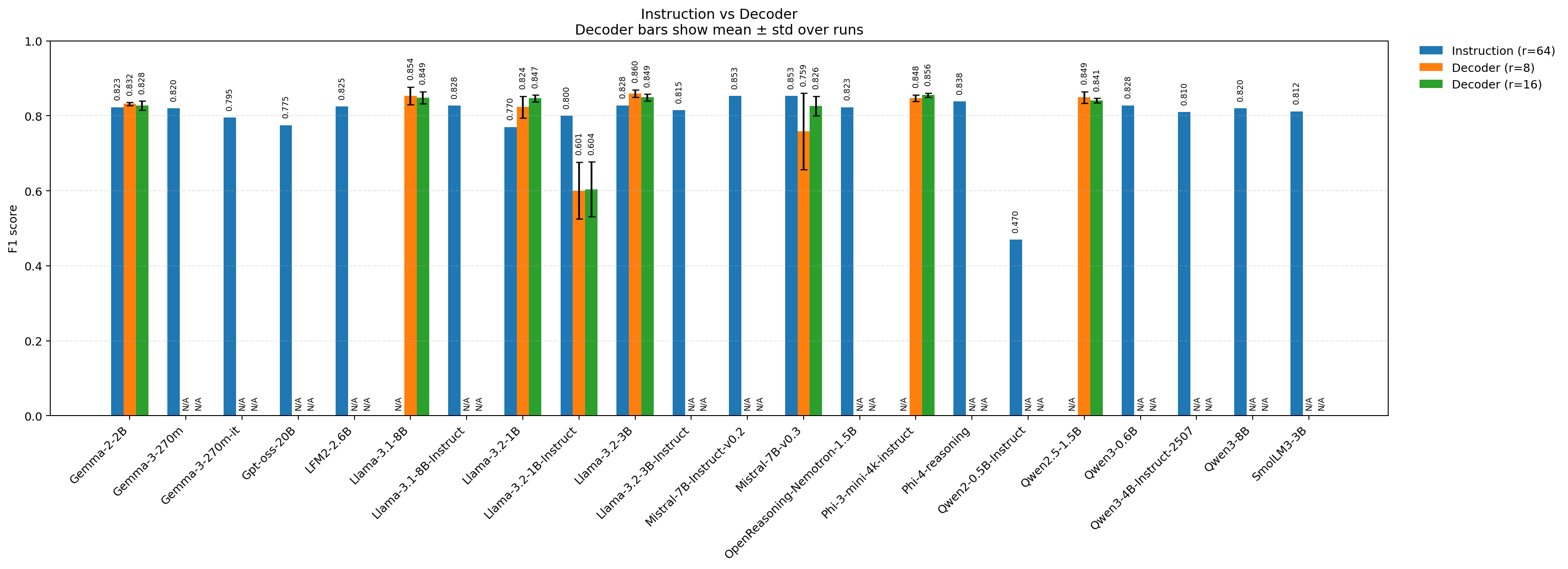}
  \caption{F1-Micro by model for the \textbf{DatasetCLV} dataset. Variability is summarized using the sample mean and standard deviation over 4 random seeds; error bars denote mean $\pm$ one standard deviation.}
  \label{fig:f1_by_model_dataset_decoder_compar2}
\end{figure*}

\begin{figure*}[t]
  \centering
  \includegraphics[width=0.85\textwidth]{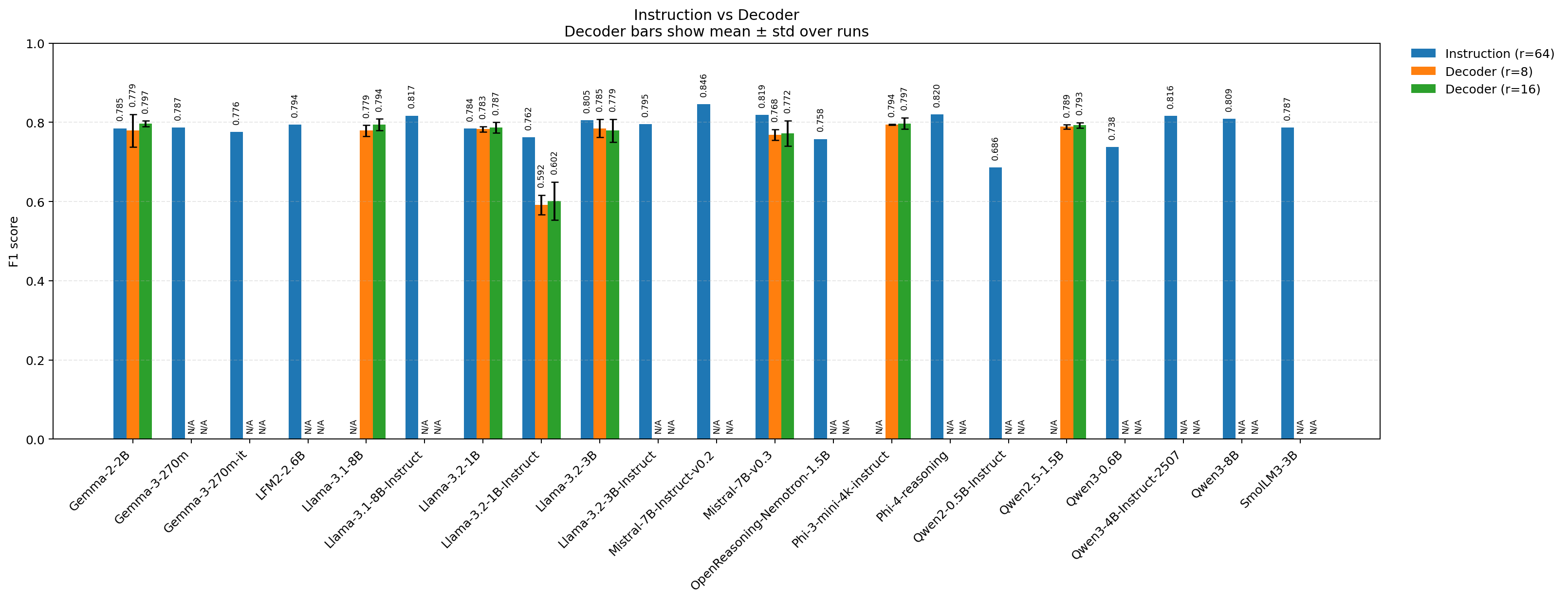}
  \caption{Micro-F1 by model for the \textbf{WIPO} dataset. Variability is summarized using the sample mean and standard deviation over 4 random seeds; error bars denote mean $\pm$ one standard deviation.}
  \label{fig:f1_by_model_dataset_decoder_compar}
\end{figure*}

Figures~\ref{fig:f1_by_model_dataset_decoder_compar2} and~\ref{fig:f1_by_model_dataset_decoder_compar} show micro-F1 broken out by model for the DatasetCLV (single-label) and WIPO (multi-label) datasets respectively. Both plots aggregate the same 4 seeds per (model, dataset) cell underlying Table~\ref{tab:direct_comparison}, with error bars denoting $\pm$ one standard deviation. The CLV chart shows tight clustering of well-converged A1 cells around 0.83--0.86 with most A2 cells in the same band; the WIPO chart shows the wider model spread and A2's reliance on larger trainable budgets.

\subsection{Full Throughput Results}
\label{sec:throughput_main}
Table~\ref{tab:sps_from_aggregates_all} extends the headline throughput summary (Table~\ref{tab:sps_from_aggregates}) to all evaluated models, including the full A2 grid and all BERT baselines. The decisive throughput advantage of BERT encoders is reproduced consistently: ModernBERT-base inference reaches $\sim$93 samples/s on CLV and $\sim$84 on WIPO, against $\sim$1--5 sps for the LLM rows. Within the LLM rows, smaller models (Llama-3.2-1B, Gemma-3-270m) reach $\sim$3--5 inference sps; larger 7--8B models stay below 1 sps inference. This throughput gap is the primary motivation for the distillation experiment in Section~\ref{sec:distillation_main}.

\begin{table*}[!t]
\footnotesize
\begin{tabularx}{\textwidth}{l X c c c c}
\toprule
 & & \multicolumn{2}{c}{\textbf{CLV (sps)}} & \multicolumn{2}{c}{\textbf{WIPO (sps)}} \\
\cmidrule(lr){3-4} \cmidrule(lr){5-6}
\textbf{Approach} & \textbf{Model} & \textbf{Train} & \textbf{Infer} & \textbf{Train} & \textbf{Infer} \\
\midrule

\multirow{8}{*}{A1 ($r$=8)}
 & Qwen2.5-1.5B & 4.91 & 3.27 & 3.84 & 2.54 \\
 & Gemma-2-2B & 4.65 & 2.29 & 3.18 & 1.85 \\
 & Llama-3.1-8B & 1.52 & 0.81 & 0.99 & 0.61 \\
 & Llama-3.2-1B & 11.37 & 4.54 & 5.51 & 3.58 \\
 & L3.2-1B-Instruct & 12.44 & 4.58 & 5.36 & 3.57 \\
 & Llama-3.2-3B & 3.04 & 1.88 & 2.05 & 1.37 \\
 & Phi-3-mini-4k-ins & 1.96 & 1.35 & 1.62 & 1.09 \\
 & Mistral-7B-v0.3 & 1.34 & 0.73 & 1.31 & 0.59 \\
\midrule

\multirow{8}{*}{A1 ($r$=16)}
 & Qwen2.5-1.5B & 6.60 & 3.29 & 4.03 & 2.55 \\
 & Gemma-2-2B & 5.13 & 2.30 & 2.11 & 1.84 \\
 & Llama-3.1-8B & 1.67 & 0.79 & 0.88 & 0.61 \\
 & Llama-3.2-1B & 9.84 & 4.56 & 5.02 & 3.56 \\
 & L3.2-1B-Instruct & 10.52 & 4.56 & 7.06 & 3.61 \\
 & Llama-3.2-3B & 2.50 & 1.87 & 1.96 & 1.37 \\
 & Phi-3-mini-4k-ins & 1.93 & 1.35 & 1.66 & 1.08 \\
 & Mistral-7B-v0.3 & 1.60 & 0.74 & 0.99 & 0.59 \\
\midrule

\multirow{10}{*}{A2 (Inst.)}
 & SmolLM3-3B & 0.96 & 1.47 & 0.80 & 0.96 \\
 & LFM2-2.6B & 1.10 & 1.69 & 0.94 & 1.26 \\
 & Qwen2-0.5B-Instruct & 2.23 & 2.97 & 2.14 & 1.58 \\
 & Qwen3-0.6B & 1.88 & 2.50 & 1.80 & 1.56 \\
 & Qwen3-4B-Instruct-2507 & 0.77 & 1.26 & 0.64 & 0.83 \\
 & Qwen3-8B & 0.46 & 0.90 & 0.39 & 0.64 \\
 & Gemma-2-2B & 1.32 & 1.86 & 1.18 & 1.37 \\
 & Gemma-3-270m & 3.50 & 5.15 & 3.25 & 3.74 \\
 & Gemma-3-270m-it & 3.48 & 5.12 & 3.21 & 3.47 \\
 & Llama-3.1-8B-Instruct & 0.47 & 0.95 & 0.41 & 0.70 \\
 & Llama-3.2-1B & 2.49 & 3.54 & 2.13 & 2.34 \\
 & Llama-3.2-1B-Instruct & 2.49 & 3.57 & 2.08 & 2.25 \\
 & Llama-3.2-3B & 1.05 & 1.69 & 0.91 & 1.16 \\
 & Llama-3.2-3B-Instruct & 1.05 & 1.72 & 0.89 & 1.15 \\
 & Phi-4-reasoning & 0.25 & 0.59 & 0.22 & 0.47 \\
 & Mistral-7B-Instruct-v0.2 & 0.45 & 0.25 & 0.39 & 0.62 \\
 & Mistral-7B-v0.3 & 0.45 & 0.23 & 0.39 & 0.62 \\
 & OpenReasoning-Nemotron-1.5B & 1.70 & 2.21 & 1.43 & 1.42 \\
 & GPT-oss-20B & 0.49 & 0.59 & -- & -- \\
\midrule

\multirow{4}{*}{BERT Baselines}
 & PatentBERT & 8.87 & 35.51 & 8.65 & 36.09 \\
 & ModernBERT-base-PT & 23.33 & 90.99 & 21.11 & 83.68 \\
 & ModernBERT-base-VX & 23.50 & 91.87 & 20.85 & 84.06 \\
 & ModernBERT-base & 24.55 & 93.45 & 21.51 & 84.48 \\
\bottomrule
\end{tabularx}
\caption{Training and inference throughput (samples per second) for CLV and WIPO datasets. A1 denotes embedding-based fine-tuning (averaged over 4 seeds); A2 denotes instruction fine-tuning.}
\label{tab:sps_from_aggregates_all}
\end{table*}

\subsection{Per-Class WIPO F1 (E8)}
\label{sec:e8_main}
WIPO's 14 categories are imbalanced; aggregate micro-F1 hides systematic differences between approaches. We report per-class precision/recall/F1 for A1 best (Llama-3.2-3B, $r$=8), A2 best (Mistral-7B-Inst-v0.2), and PatentBERT in the released artifact (\texttt{paper\_data/E8/per\_class\_f1.json} and the corresponding heatmap PDF). A1 wins on the head classes (Electrical Engineering, Mechanical Engineering, Chemistry) by 1--3 F1 points; A2 wins on a small number of tail classes where its larger trainable budget appears to help generalization to rare label combinations.

\section{Conclusion}
This study systematically compared two strategies for adapting large decoder-only LLMs to patent text classification: an embedding-based approach using a lightweight classification head and an instruction-based approach leveraging \textit{prompt-response} supervised generation, both fine-tuned via LoRA and quantization. Across both single-label and multi-label patent datasets, the embedding head matches or exceeds instruction tuning and the BERT baselines at $10$--$30\times$ fewer trainable parameters; instruction tuning catches up only on multi-label WIPO and only at $\geq$167M trainable parameters (comparable to the \emph{total} parameter count of our BERT baselines), so the ``parameter-efficient'' framing holds cleanly for the single-label regime and is qualified for multi-label. BERT encoders retain a throughput advantage, but this gap is largely recoverable: distilling our Approach-1 Llama-3.2-3B teacher into a ModernBERT-base student retains $\geq$97\% of teacher F1 (0.845 CLV, 0.851 WIPO; Table~\ref{tab:e6_kd}, Section~\ref{sec:distillation_main}) at $\sim$45$\times$ higher inference throughput.

Beyond accuracy, the embedding head's outputs are empirically calibrated (ECE~$=$~0.094, Brier~$=$~0.232 on CLV; ECE~$=$~0.017, Brier~$=$~0.019 on WIPO per-label sigmoid; full reliability diagrams in Section~\ref{sec:calibration_main}), in contrast to the format fragility and parsing brittleness reported for instruction-tuned classifiers \citep{ghosh2024closer} and reproduced by our verbalizer ablation (Section~\ref{sec:verbalizer_main}, where A2's WIPO F1 swings from 0.262 to 0.838 across verbalizer choices). Together with the distillation result above, this positions embedding-head adaptation of causal LLMs as a practical default for domain text classification under single-GPU constraints --- delivering BERT-class accuracy and calibration today, with a clear path to BERT-class throughput via distillation.

\section*{Limitations}
We surface six limitations that bound the generality of our claims.

\paragraph{Proprietary single-label data.} Half of our headline numbers rest on an internal corpus (DatasetCLV) that cannot be redistributed; external replication is therefore restricted to the WIPO subset, where our A1-vs-A2 comparison shows a smaller and partially inverted gap.

\paragraph{Single-domain scope.} Headline numbers are on patent text; we externally validate on AG~News (Section~\ref{sec:e2_main}) but do not extend to multi-domain hierarchical benchmarks such as EUR-Lex, and patent-specific pre-training of the BERT baselines confounds the LLM-vs-BERT comparison.

\paragraph{Significance gap is small but not certified.} We report paired McNemar tests and bootstrap $\Delta$F1 95\% CIs for the A1-vs-A2 contrast (Section~\ref{sec:e1_main}); neither contrast reaches $p{<}0.05$ on per-instance correctness ($p$=0.24 CLV, $p$=0.51 WIPO) and both 95\% CIs include zero. The numerical advantage of A1 is consistent in direction but not statistically certified.

\paragraph{Throughput penalty.} Approach~1 is $\sim$20$\times$ slower at inference than ModernBERT-base at parity F1 (Table~\ref{tab:sps_from_aggregates}); distillation (Section~\ref{sec:distillation_main}) closes most of this gap at a $\leq$1.5-point F1 cost, but for the highest-throughput deployments BERT-class encoders remain preferable.

\paragraph{Best multi-label configuration is not parameter-efficient.} Mistral-7B with Approach~2 reaches F1~0.819 on WIPO using 167.8M trainable parameters --- comparable to the \emph{total} parameter count of our BERT baselines --- so the ``parameter-efficient'' framing holds for the single-label regime but is qualified for multi-label.

\paragraph{Limited mechanistic analysis.} Calibration is now quantified empirically (ECE/Brier in Section~\ref{sec:calibration_main}); however, representation-geometry analyses (e.g., t-SNE/UMAP, CKA) that would mechanistically explain Approach~1's single-label advantage are deferred to future work.

\paragraph{Training instability across seeds.} A subset of seeds for the A1 Llama-3.2-3B baseline (3 of 4 WIPO seeds and 1 of 4 CLV seeds) and 1 of 4 seeds in the max-pooling ablation cell failed to converge during fine-tuning despite identical hyperparameters and code path, predicting near-random or majority/empty labels (test F1~$<$~0.5). We report results from the remaining converged seeds and flag the affected cells in Table~\ref{tab:direct_comparison}; replicators should expect to occasionally see this failure mode and we recommend running each cell with 6+ seeds and reporting only the converged subset.

\section{Future Work}
Future work includes exploring hybrid fine-tuning pipelines (combining embedding and instruction paradigms), scaling to even larger models and more challenging multi-label or hierarchical datasets, and conducting in-depth representation analyses to further understand the internal dynamics that drive the success of embedding-based tuning. Extensions to regression and other structured tasks also offer promising directions.

We also would like to further investigate the potential of reasoning models for classification tasks. In this work, we reported results using OpenReasoning-Nemotron-1.5B with instruction tuning but the standard prompt we used may not be optimal for fully leveraging thinking models like this one.

An interesting direction for future work is to compare the embeddings produced by decoder-based and BERT-based models using t-SNE or UMAP, and to quantify class separability via cluster-quality metrics (e.g., silhouette, Davies–Bouldin) and representation-similarity analyses (e.g., CKA). These analyses would indicate how well each model separates classes in the embedding space and which yields more distinct, interpretable clusters. In parallel, because decoder-only LLMs are substantially larger and exhibit lower training/inference throughput than encoder-only BERT-style models in this setting, a teacher–student (knowledge distillation) approach is a natural avenue: treat the decoder model as a teacher and train a compact BERT-style student with temperature-scaled soft targets and, where feasible, intermediate feature/attention matching. For hierarchical, multi-label settings, sequence-level and contrastive distillation objectives aligned with the label taxonomy are promising directions. The objective is to retain most of the teacher's accuracy while recovering the throughput and memory efficiency characteristic of BERT-style models.

\paragraph{Remaining empirical work.} The companion experiments planned to tighten the conclusions of this study --- paired significance tests, a third public benchmark, LLM2Vec / PatentSBERTa baselines, pooling-strategy ablation, calibration, prompting baselines, verbalizer ablation, per-class WIPO breakdown, and a knowledge-distillation throughput recovery --- are now reported in Sections~\ref{sec:e1_main}--\ref{sec:e8_main} and summarized in the main body. The principal remaining open direction is \textit{representation-geometry analysis}: comparing decoder-vs-encoder embeddings via t-SNE/UMAP and cluster-quality metrics (silhouette, Davies--Bouldin) and via CKA / representational-similarity analyses, to mechanistically explain Approach~1's single-label advantage.

\section{Practical Implications for Practitioners: Robust Fine-Tuning of LLMs for Classification}
\label{sec:practical}
Partial ablation studies and error analyses offer several practical takeaways for engineers and practitioners seeking reliable, efficient LLM pipelines for text classification tasks. While resource constraints prevented exhaustive ablations across all models and hyperparameters, the experiments conducted illuminate key design choices that have direct consequences on deployment, scalability, and model robustness.

\paragraph{LoRA and Lightweight PEFT Are Essential:} Allowing the model to adapt internal representations with LoRA adapters, even with a very small number of trainable parameters (e.g., 8M–24M out of billions), produced a substantial boost in F1 over “frozen” LLMs where only the classifier head is fine-tuned. This confirms that adapting deeper layers, not just leveraging static pretrained embeddings, is necessary for strong performance—especially on complex or domain-specific tasks.

\paragraph{Adapter Rank and Tuning:}
We found that increasing the LoRA rank from 8 to 16 provides a small but stable benefit, especially for highly multi-label tasks (like WIPO). Lower ranks may suffice for simpler datasets or where compute is the tightest bottleneck, but practitioners should not hesitate to test slightly larger LoRA ranks if throughput and memory permit, as the resource tradeoff remains minimal compared to classical full-model fine-tuning.

\paragraph{Supervised Adaptation Outperforms Frontier Prompting:}
We empirically test whether headless prompting matches supervised LoRA adaptation by evaluating four frontier API models (GPT-5, GPT-5-mini, GPT-5-nano, Claude Opus 4.6) prompted with our verbatim A2 templates across zero-shot, 5-shot, and chain-of-thought modes (24 cells; Section~\ref{sec:e3d_main}). The best closed-source result reaches 0.823 micro-F1 on CLV and 0.752 on WIPO --- versus 0.860 and 0.785 for our 1--3B LoRA Approach-1 model (4-seed campaign values; Table~\ref{tab:direct_comparison}), a 3.3--3.7 point gap. Smaller frontier tiers fall further behind, and chain-of-thought reasoning typically underperforms zero-shot on these strict-format classification tasks. Genuine supervised adaptation --- not prompt engineering, even with state-of-the-art frontier models --- is required for best-in-class accuracy in this regime.

\paragraph{Output Structure (by construction):}
The embedding-based approach emits a calibrated $C$-way softmax (or sigmoid) probability vector, which admits straightforward thresholding and confidence-based routing without any post-processing. The instruction-based approach emits a generated text sequence that requires constrained decoding or string parsing to extract a label. This is a structural difference, and is now also empirically supported: A1 (Llama-3.2-3B, seed~1) achieves ECE=0.094 and Brier=0.232 on CLV, with reliability diagrams in Section~\ref{sec:calibration_main}.

\paragraph{Error Sources and Engineering Overheads:}
Instruction-based models were more brittle with respect to small changes in prompt phrasing or formatting, leading to label extraction errors and reduced performance. When models did not reproduce the expected verbalization template exactly, result parsing became nontrivial. Engineers deploying such models should budget time for extra evaluation and safeguards if instruction-tuned models are used in production.

\paragraph{Resource-Efficient Choices:}
Overall, resource-lean, quantized LLMs with LoRA adapters offer a substantial reduction in training and inference latency—enabling rapid prototyping and frequent updates even on moderate hardware. Embedding-based designs proved “plug-and-play”—easy to implement, robust to input variations, and requiring less custom post-processing, which is critical for maintaining software simplicity and reducing operational costs.

\section{Hyperparameters}
\label{sec:hyperparams}

\begin{table*}[t] 
\small 
\begin{tabularx}{\textwidth}{lXX} 
\toprule
\textbf{Hyperparameter} & \textbf{Instruction Fine-Tuning} & \textbf{Decoder Tuning} \\
\midrule
Task formulation & N/A & Single-label (5 classes) \\
Fine-tuning regime & QLoRA (4-bit base) & QLoRA (4-bit base) \\
Weight quantization & 4-bit (\texttt{bitsandbytes}) & 4-bit (\texttt{bitsandbytes}) \\
Compute precision & \texttt{bfloat16} (mixed) & \texttt{bfloat16} (mixed) \\
LoRA rank $r$ & 64 & 8, 16 \\
LoRA $\alpha$ & 16 & 16 \\
LoRA dropout & 0.05 & 0.05 \\
LoRA bias & none & none \\
LoRA target layers & All 4-bit linear (excl. \texttt{lm\_head}) & N/A \\
Tokenizer settings & N/A & \texttt{prefix\_space=T}; \texttt{pad=eos\_id} \\
Padding side & Right; \texttt{pad=eos\_id} & Dynamic (\texttt{DataCollator}) \\
Max seq length & N/A & 1024 (truncate) \\
Optimizer & Paged AdamW (32-bit) & AdamW (8-bit) \\
Learning rate & $2\times10^{-4}$ & $2\times10^{-4}$ \\
Weight decay & $1\times10^{-3}$ & 0.01 \\
Max gradient norm & 0.3 & 1.0 \\
LR scheduler & Cosine & Linear; no warmup \\
Warmup ratio & 0.03 & N/A \\
Training epochs & 5 & 20 (max) \\
Early stopping & N/A & Patience = 2 epochs \\
Batch size (per dev) & 1 & 1 \\
Eval batch (per dev) & 1 & 4 \\
Effective batch size & $1 \times 8 \times n_{\text{devices}}$ & $1 \times 8 \times n_{\text{devices}}$ \\
Grad checkpointing & Enabled (non-reentrant) & N/A \\
Model selection & N/A & Best (micro-F1) \\
Evaluation / saving & N/A & Every epoch; \texttt{limit=1} \\
Random seed & 42 & Exp-dependent \\
\bottomrule
\end{tabularx}
\caption{Comparison of training hyperparameters and settings for instruction fine-tuning (Approach~2) and embedding-head decoder tuning (Approach~1). Optimizer differs by approach (Paged AdamW 32-bit for A2 vs.\ AdamW 8-bit for A1) due to the larger trainable budget A2 requires; both are widely used variants and the asymmetry is acknowledged as a minor confounder for direct head-to-head comparison. A1 uses a fixed random seed=42 across the 4-seed re-runs; A2's seed is experiment-dependent (per the original training campaign).}
\label{tab:hyperparams_combined}
\end{table*}

\bibliographystyle{unsrtnat}
\bibliography{references}

\appendix

\section{Dataset Information}
\begin{table}[b]
\centering
\setlength{\tabcolsep}{5pt} 
\begin{tabular}{lrrrr}
\toprule
\textbf{Dataset} & \textbf{Labels} & \textbf{Train} & \textbf{Val.} & \textbf{Test} \\
\midrule
WIPO       & 14 & 1731  & 424  & 533  \\
DatasetCLV &  5 & 1481  & 371  & 400  \\
\bottomrule
\end{tabular}
\caption{Dataset statistics for downstream evaluation tasks.}
\label{tab:datasets}
\end{table}
\label{app:datainfo}
WIPO Vision Dataset (WIPO) \\
\textbf{Categories:}
\begin{itemize}
\item Adaptive Focus
\item Artificial Iris
\item Artificial Silicon Retina (ASR) / Retinal Prostheses
\item Augmented Reality Devices
\item Bionic Eye (System)
\item Cortical Implants
\item Drug Delivery (Vision-related)
\item Hand Wearables
\item Intraocular Lenses (IOL) with Sensors
\item Intracorneal Lenses
\item Multifocal
\item Smart Eyewear
\item Telescopic Lenses
\item Virtual Reality Devices\\
\end{itemize}
\begin{flushleft}
DatasetCLV\\
\end{flushleft}
\textbf{Categories:} Five categories related to data storage and networking.

\subsection*{Data Statement}
Following \citet{bender2018data}, we provide a structured data statement for each of our datasets.

\paragraph{WIPO-Alpha (Vision subset).} A publicly available subset of the World Intellectual Property Organization patent classification corpus, distributed by WIPO for research on hierarchical patent categorization (English). \emph{Curation rationale:} the dataset is constructed from real patent filings to evaluate hierarchical multi-label classification along the WIPO classification taxonomy. \emph{Language variety:} English, professional/technical register. \emph{Speaker/author demographics:} not available --- patent filings are corporate or attorney-drafted and the dataset does not include author metadata. \emph{Annotator demographics:} labels are derived from the existing WIPO classification system applied by patent examiners. \emph{Speech situation:} formal written technical text. \emph{Text characteristics:} technical, professionally drafted, contains domain-specific terminology; average length and class-distribution statistics are in Table~\ref{tab:datasets}. \emph{Class imbalance:} the 14 WIPO sections are skewed; head classes (Electrical Engineering, Chemistry, Mechanical Engineering) account for substantially more documents than tail classes (Astronautics, Agriculture-related sub-codes); per-class counts are in Table~\ref{tab:datasets}. \emph{License and use restrictions:} the WIPO-Alpha collection is released by WIPO for non-commercial research and educational use; redistribution is permitted under that scope. We do not redistribute the data with this work but provide references to the official WIPO source. No personally identifying or sensitive information is associated with the patent records.

\paragraph{DatasetCLV (proprietary, single-label).} An internal English-language corpus of 2{,}252 patent documents labeled across five technical categories related to networked-storage products, drawn from an operational patent-classification workflow. \emph{Curation rationale:} the dataset was built to support an internal categorization product; labels come from a fixed five-class taxonomy that was operationally relevant at the time of curation. \emph{Language variety:} English. \emph{Annotator demographics:} labels were assigned by domain experts in the course of normal business operations rather than collected for this study; no demographic data was retained. \emph{Class imbalance:} the five-class distribution is moderately skewed but not extreme; per-class counts are in Table~\ref{tab:datasets}. \emph{Sensitive content:} the corpus contains no personally identifying information; all text is drawn from publicly granted patent filings. \emph{License and use restrictions:} the source data is licensed for internal use by the data owner; the curated dataset cannot be redistributed under that license. We report aggregate statistics, hyperparameter settings, and per-class counts to support partial replicability and have released public-data results (WIPO-Alpha, AG~News) for external comparison.

\section{Prompt Templates}
\label{sec:prompt-templates}
\subsection{Single-Label}
\subsubsection{Training prompt template (with gold answer)}

\noindent We use the following template during training, where
\verb|<CHOICE_SET>| is a space-separated list of identifier--label pairs
(e.g., \verb|A:Label1 B:Label2 C:Label3|), \verb|<TEXT>| is the input
instance, and \verb|<GOLD_ID>| / \verb|<GOLD_LABEL_NAME>| are the gold
identifier and label name:

\begin{tcolorbox}[
    colback=gray!5, 
    colframe=gray!50, 
    size=small, 
    arc=0mm, 
    left=2pt, right=2pt,
    boxrule=0.5pt,
    fontupper=\small\ttfamily,
    breakable
]
You are a classifier. Choose exactly one label from the set below. \\
Choices: <CHOICE\_SET> \\
Output format: <ID>\textbackslash t<LABEL\_NAME> \\
Do not add anything else.

\medskip
\textbf{TEXT:} \\
<TEXT>

\medskip
\textbf{ANSWER:} \\
<GOLD\_ID> \quad <GOLD\_LABEL\_NAME>
\end{tcolorbox}

\subsubsection{Test prompt template (no answer filled in)}

\noindent At inference time, we remove the gold answer and leave the
\verb|ANSWER:| field empty:

\begin{tcolorbox}[
    colback=gray!5,
    colframe=gray!80,
    sharp corners,
    boxrule=0.5pt,
    fontupper=\small\ttfamily,
    width=\columnwidth,
    breakable
]
You are a classifier. Choose exactly one label from the set below. \\
Choices: <CHOICE\_SET> \\
Output format: <ID>\textbackslash t<LABEL\_NAME> \\
Do not add anything else.

\medskip
\textbf{TEXT:} \\
<TEXT>

\medskip
\textbf{ANSWER:}
\end{tcolorbox}

\subsection{Multi-Label}
\subsubsection{Multi-label training prompt template}

\noindent For multi-label classification, let \texttt{ALLOWED} denote the
ordered list of allowed labels and \texttt{<ALLOWED\_LABEL\_SET>} be the
string representation shown to the model (e.g., a comma- or newline-separated
list of labels). \texttt{<FIRST\_ALLOWED\_LABEL>} denotes the first label in
\texttt{ALLOWED}, \texttt{<TEXT>} is the input text instance, and
\texttt{<GOLD\_LABELS\_JSON>} is a JSON-style list of the gold label names.
The training-time prompt template is:

\begin{tcolorbox}[
    colback=gray!5,
    colframe=gray!80,
    sharp corners,
    boxrule=0.5pt,
    fontupper=\small\ttfamily,
    width=\columnwidth,
    breakable
]
You are a careful multi-label classifier. Choose zero or more labels from this allowed set: \\
<ALLOWED\_LABEL\_SET>

\medskip
Return STRICTLY a JSON list under the key "labels" and nothing else, e.g.: \\
labels: ["<FIRST\_ALLOWED\_LABEL>"] or labels: []

\medskip
\textbf{text:} <TEXT> \\
\textbf{labels:} <GOLD\_LABELS\_JSON>
\end{tcolorbox}

\subsubsection{Multi-label test prompt template}

\noindent At test time, we use the same instruction and allowed label set,
but leave the value after \texttt{labels:} empty for the model to fill:

\begin{tcolorbox}[
    colback=gray!5,
    colframe=gray!80,
    sharp corners,
    boxrule=0.5pt,
    fontupper=\small\ttfamily,
    width=\columnwidth,
    breakable
]
You are a careful multi-label classifier. Choose zero or more labels from this allowed set: \\
<ALLOWED\_LABEL\_SET>

\medskip
Return STRICTLY a JSON list under the key "labels" and nothing else, e.g.: \\
labels: ["<FIRST\_ALLOWED\_LABEL>"] or labels: []

\medskip
\textbf{text:} <TEXT> \\
\textbf{labels:}
\end{tcolorbox}

\section{Per-Model F1 Results with 95\% CIs}
\label{sec:appendix_results}
\begin{figure*}[t]
  \centering
  \includegraphics[width=0.8\textwidth]{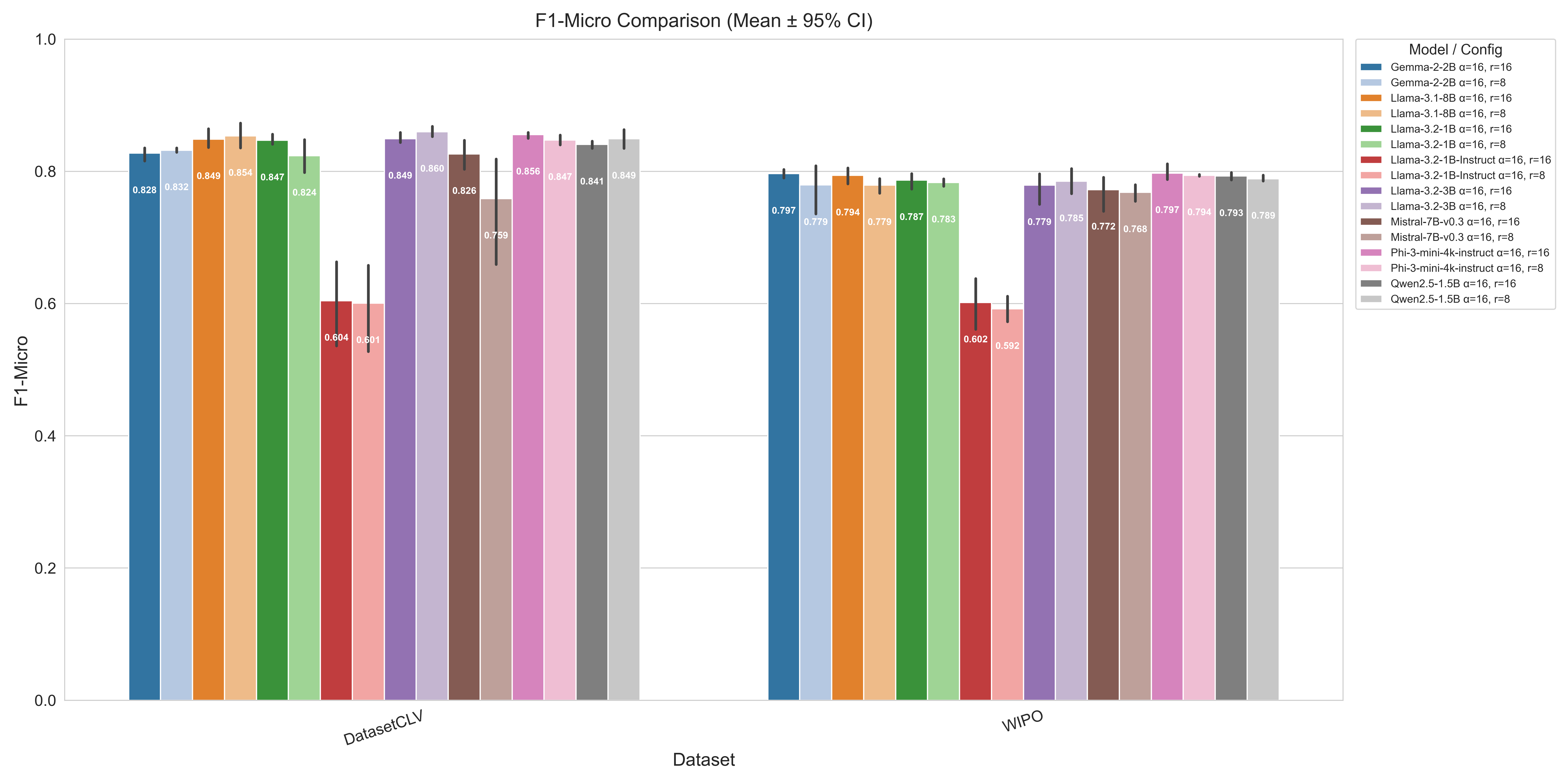}
  \caption{F1-Micro by model and dataset using decoder tuning. Error bars show 95\% confidence interval computed over \textbf{4 random seeds} per (dataset, model). Confidence intervals are estimated using the Student's $t$ distribution and clipped to $[0,1]$.}
  \label{fig:f1_by_model_dataset_decoder}
\end{figure*}

\begin{figure*}[t]
  \centering
  \includegraphics[width=0.8\textwidth]{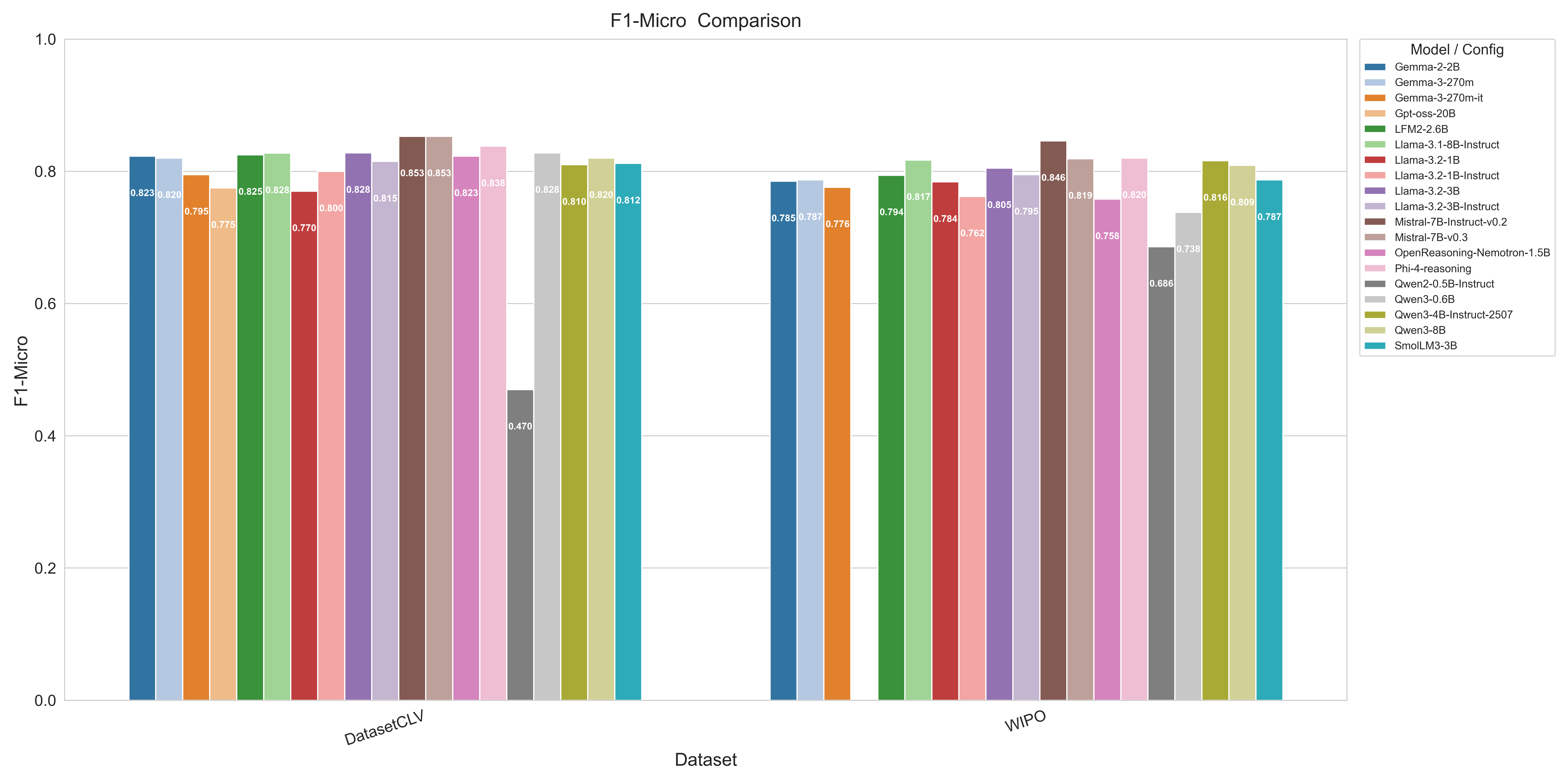}
  \caption{F1-Micro by model and dataset using instruction tuning.}
  \label{fig:f1_by_model_dataset_instruct}
\end{figure*}

\FloatBarrier

\end{document}